\documentclass[a4paper, 11pt]{article}

\usepackage[utf8]{inputenc}
\usepackage{amsmath}
\usepackage{amsfonts,amssymb,amsopn,amscd,amsthm}
\usepackage{stix2}
\usepackage{comment}
\usepackage{subcaption}
\usepackage{color}
\usepackage{xcolor}
\usepackage[colorlinks]{hyperref}
\usepackage[left=3.5cm,top=2.5cm,bottom=2.5cm,right=3cm]{geometry}
\usepackage{mathtools, stmaryrd}
\usepackage{bm}
\usepackage{authblk}

\usepackage[parfill]{parskip}

\usepackage[numbers]{natbib}

\date{\today}

\allowdisplaybreaks[4]

\DeclareRobustCommand{\SkipTocEntry}[5]{}

\DeclareMathOperator*{\argmin}{argmin}

\DeclareMathOperator{\UDF}{\textrm{UDF}}
\DeclareMathOperator{\SDF}{\textrm{SDF}}

\DeclareMathOperator{\AD}{\textrm{AD}}

\def\1{{\mathbf 1}}

\def\R{{\mathbb R}}

\def\P{{\mathbb P}}

\def\Bc{{\mathcal B}}

\def\Fc{{\mathcal F}}

\def\Ic{{\mathcal I}}

\def\Lc{{\mathcal L}}

\def\Nc{{\mathcal N}}
\def\Oc{{\mathcal O}}

\def\Rc{{\mathcal R}}

\def\Uc{{\mathcal U}}
\def\Vc{{\mathcal V}}

\def\bb{\bm{b}}
\def\xb{\bm{x}}

\def\zb{\bm{z}}

\def\thetab{\bm{\theta}}
\def\Bb{\bm{B}}

\def\Xb{\bm{X}}

\newtheoremstyle{break}
  {\topsep}{\topsep}%
  {\itshape}{}%
  {\bfseries}{}%
  {\newline}{}%
\theoremstyle{break}

\newtheorem{thm}{Theorem}[section]

\newtheorem{definition}[thm]{Definition}

\DeclarePairedDelimiterX\Set[1]{\lbrace}{\rbrace}%
 {  #1 }

\hypersetup{%
    pdfborder = {0 0 0},
    colorlinks,
    citecolor=green,
    filecolor=Darkgreen,
    linkcolor=blue,
    urlcolor=cyan!50!black!90
}

\providecommand{\keywords}[1]{\textbf{\textit{Keywords:}} #1}

\title{Statistical Edge Detection And UDF Learning For Shape Representation}

\author[1]{Virgile Foy \thanks{virgile.foy@math.univ-toulouse.fr}}
\author[1]{Fabrice Gamboa \thanks{https://www.math.univ-toulouse.fr/~gamboa/}}
\author[1]{Reda Chhaibi \thanks{https://www.math.univ-toulouse.fr/~rchhaibi/}}
\affil[1]{Institut de Mathématiques de Toulouse, Université Paul Sabatier\\
  118 Route de Narbonne\\
  Toulouse, 31400, France}

\begin{document}

\maketitle

\begin{abstract}

In the field of computer vision, the numerical encoding of 3D surfaces is crucial. It is classical to represent surfaces with their Signed Distance Functions (SDFs) or Unsigned Distance Functions (UDFs) \cite{atzmon2020sal, chen2019learning, gu2020cascade, mescheder2019occupancy}. For tasks like representation learning, surface classification, or surface reconstruction, this function can be learned by a neural network, called Neural Distance Function \cite{chibane2020neural}. This network, and in particular its weights, may serve as a parametric and implicit representation for the surface. The network must represent the surface as accurately as possible. In this paper, we propose a method for learning UDFs that improves the fidelity of the obtained Neural UDF to the original 3D surface. The key idea of our method is to concentrate the learning effort of the Neural UDF on surface edges.  More precisely, we show that sampling more training points around surface edges allows better local accuracy of the trained Neural UDF, and thus improves the global expressiveness of the Neural UDF in terms of Hausdorff distance. 

To detect surface edges, we propose a new statistical method based on the calculation of a $p$-value at each point on the surface. Our method is shown to detect surface edges more accurately than a commonly used local geometric descriptor \cite{pauly2002efficient}.

\end{abstract}

\keywords{Computer Vision, Edge Detection, Goodness-Of-Fit Tests, Neural Networks, Representation Learning}

\newpage 

\tableofcontents

\newpage


\section{Introduction}

\subsection{Industrial Context}
Many problems in science and engineering require solving complex boundary value problems. The geometric shape of one part of a system may have a significant influence on the system's performance. For example, the geometric shape of the turbine blades of an aircraft engine significantly influence the engine's performance \cite{rollsroyce1996}. In those cases, the geometry of the parts must be determined very precisely. The simulation codes used to design the parts can be very costly in terms of time and power \cite{casenave2024mmgp}. The ultimate goal of our work is to obtain real-time estimates of the outputs of these codes. This would significantly reduce the design time and allow better exploration of different geometries for potentially higher performance. Such approach has also motivated the seminal works \cite{remelli2020meshsdf} and \cite{park2019deepsdf} that have inspired our research.

In order to build such a model, we isolate the part from the industrial system and consider it as a shape in $\R^3$. We have access to a set of these shapes and we attempt to construct a parametric and implicit representation of these shapes. Parametric in the sense that each shape from the dataset is mapped to a combination of parameters called latent representation. Implicit means that the shapes are numerically encoded by their distance functions, from which they are the zero-level set (see Section \ref{subsection:intro_implicit_encoding}). Practically, each shape is represented by the weights of a neural network called Neural Unsigned Distance Function (Neural UDF, see Sections \ref{subsection:neural_udf_learning} and \ref{subsection:train_neural_udf}). Each Neural UDF represents each shape because it has been trained to reproduce the shape's true UDF. This latent representation can be used as the input data of a regression model to predict numerical simulations outputs.

In this paper, we only focus on the problem of learning an UDF by a neural network. In particular, we propose to train the Neural UDF on points sampled in a smart way. Our objective is that the Neural UDF represents each shape more accurately. 
	
	\subsection{Encoding 3D Shapes With Implicit Field}
        \label{subsection:intro_implicit_encoding}
	
Some active research fields like computer vision, robotics, and numerical simulations require the ability to manipulate three-dimensional data. The 3D scenes to be processed consist of sets of 3D objects. Here we focus on the envelopes of these objects, specifically 3D watertight surfaces (see Definition \ref{def:watertight_surface}).

\begin{definition}[Watertight surface]
\label{def:watertight_surface}

Let $\Vc$ be a subset of $\R^3$ and $\Bc$ its boundary. $\Bc$ is a \textbf{watertight surface} \textit{i.i.f.} the three following conditions are fulfilled:
\begin{itemize}
\item $\Vc$ is bounded
\item $\Vc$ is a connected set
\item $\R^D \setminus \Vc$ is a connected set
\end{itemize}
\end{definition}

Given a task to be performed on a set of surfaces (representation learning, regression, classification,...), a crucial preliminary question is the digital format used to encode the surfaces. This format may be imposed by the data source or chosen to suit the learning approach.

The most common representation is meshing. Meshes are particularly used in CFD or design. It is also well-suited for training graph networks \cite{scarselli2009gnns, lino2021simulating, lino2022multi}. In this work, we only used surface meshes in order to sample points from the surfaces and compute the surface UDF (see Section \ref{subsection:compute_true_udf}).

Point cloud is also an important type of 3D geometric data structure. This type of data is becoming increasingly available as acquisition democratizes with stereo cameras and LiDAR. However, extracting geometric features from a point cloud is challenging because the number of sampled points and the order in which they are recorded can vary within a given set of surfaces. Many Geometric Deep Learning approaches are based on PointNet model \cite{qi2017pointnet}. This model allows to learn shape representation for regression and classification tasks. In our work, we use point cloud representation for edge detection, surface reconstruction and visualization.

Other approaches are based on voxel grids \cite{ghadai2019multi}. This format is well-suited for 3D convolutional networks but is very expensive in terms of memory. We did not use this approach here.

It is also possible to encode a 3D surface in the form of its distance function \cite{atzmon2020sal, chen2019learning, gu2020cascade, mescheder2019occupancy}. More precisely, this function associates each point in the ambient space with its distance to the closest object in the surface. This function can be signed (Signed Distance Function, SDF) or unsigned (Unsigned Distance Function, UDF). These functions are formally defined in \ref{def:sdf} and \ref{def:udf}.

\begin{definition}[Signed Distance Function]
\label{def:sdf}
Let us consider a \textit{watertight surface} $\Bc \subset \R^3$. As it is watertight, one can define $\Ic_{\Bc}$ (\textit{resp.} $\Oc_{\Bc}$) the inner (\textit{resp.} outer) part of $\Bc$. The \textbf{Signed Distance Function (SDF)} of $\mathcal{B}$ is the function $\mathrm{SDF}_{\Bc}: \R^3 \rightarrow \R$ such that:
\begin{equation}
     \forall \bm{x} \in \R^3, \ \mathrm{SDF}_{\Bc}(\bm{x}) \vcentcolon = 
     \begin{cases} 
        d(x,\Bc) \: \text{if $\bm{x}\in\Oc_{\Bc}$} \\ 
        -d(x,\Bc) \: \text{if $\bm{x}\in\Ic_{\Bc}$} 
    \end{cases}
\label{eq:def_sdf}
\end{equation}
\end{definition}

\begin{definition}[Unsigned Distance Function]
\label{def:udf}
Let $\Bc \subset \R^3$ be a \textit{watertight surface}. The \textbf{Unsigned Distance Function (UDF)} of $\mathcal{B}$ is the function $\mathrm{UDF}_{\Bc}: \R^3 \rightarrow \R$ such that:
\begin{equation}
    \forall \bm{x} \in \R^3, \ \mathrm{UDF}_{\Bc}(\bm{x}) \vcentcolon= d(\bm{x},\Bc).
\label{eq:def_udf}
\end{equation}
\end{definition}

The surface thus corresponds to the zero-level set of its distance function. We use this format because it gives a continuous representation and can be easily learned using a surrogate model (for example a neural network). In Section \ref{subsection:intro_deepsdf}, we give a pratical description of a representation learning model based on SDFs. However, the contributions of this paper relate to methods based on UDFs.
	
	    \subsection{\textit{DeepSDF}: Autodecoder For Representation Learning} 
        \label{subsection:intro_deepsdf}
	
In \cite{park2019deepsdf}, Park et. al. propose a model called \textit{DeepSDF} for learning representation of 3D watertight surfaces based on implicit distance function and auto-decoder architecture. This model combines the following two ideas:
\begin{enumerate}
\item \label{item:sdf_learning} interpretation of a surface as the zero-level set of its SDF. Train a neural network to predict the values of this function on a dataset of points in $\R^3$.
\item \label{item:representation_learning} learn a representation $\zb$ of a surface (encoded by a mesh) in a low-dimensional space, as inspired by autoencoder models \cite{ballard1987}.
\end{enumerate}

This Section gives a practical description of the methodology proposed in \cite{park2019deepsdf}. We make this description rough enough so that the main ideas can be easily understood. For more details, we refer to \cite{park2019deepsdf}.

In Paragraph \ref{subsubsection:deepsdf_pb_statement}, we formulate the problem tackled by \textit{DeepSDF} model. In Paragraph \ref{subsubsection:data_preparation}, we describe the dataset used to train the network. In Paragraph \ref{subsubsection:learning_latent_representation}, we explain how the netowrk is trained. In Paragraph \ref{subsubsection:inference}, we give the procedure used to map any new shape to a point in the latent space. In Paragraph \ref{subsubsection:deepsdf_shape_generation}, we explain how one can use DeepSDF model for shape generation.

            \subsubsection{Problem Statement}
            \label{subsubsection:deepsdf_pb_statement}

Consider a set of watertight surfaces $\Bc_1,...,\Bc_N$ whose Signed Distance Functions $\SDF_1,...,\SDF_N$ can be calculated at any point of $\R^3$. The SDF of a surface is formally defined in \ref{def:sdf}. 

\textit{DeepSDF} model consists of an auto-decoder network \cite{auto_decoder2019}. This single network is able to 
\begin{itemize}
    \item estimate the SDF of each of the shapes $\Bc_1,...,\Bc_N$ at any point in $\R^3$. 
    \item maps the shapes $\Bc_1,...,\Bc_N$ to latent codes $\bm{z_1},...,\bm{z_N} \in \R^L $. Latent dimension $L$ is a hyperparameter of the model that needs to be fixed in advance.  
\end{itemize}

\subsubsection{Data preparation}
\label{subsubsection:data_preparation}

The training dataset is composed of inputs and target ouputs. The inputs are the concatenation of:
\begin{itemize}
    \item the spatial coordinates of a point in $\R^3$
    \item a latent code in $\R^L$.
\end{itemize}
For any point $\bm{x} \in \R^3$ and any latent code $\bm{z} \in \R^L$ (refering to a shape $\Bc$), the associated target output in the dataset is the value of the SDF of $\Bc$ at the point $\bm{x}$.

The training dataset $\bm{X}$ is built as the union of $N$ sub-datasets - one for each shape:
\begin{equation}
\bm{X} = \bigcup_{j=1}^{N} \bm{X_j}.
\label{eq:X_as_an_union}
\end{equation}

Given a shape index $j\in\llbracket 1,N\llbracket$, let us describe how to build the dataset $\bm{X_j}$ corresponding to the shape $\Bc_j$. A set of points $\bm{x_j^{(1)}},...,\bm{x_j^{(T)}}$ is sampled in the ambient space $\R^3$. These points are sampled close to the surface $\Bc_j$
\footnote{For the sake of simplicity, we do not dive into details regarding the way training points are sampled as it is the main topic of Section \ref{subsection:sample_training_points}. For more information on this, we refer to the supplementary materials of \cite{park2019deepsdf}.}
. The obtained training points constitute the data sets
\begin{equation}
\bm{X_j} \vcentcolon= \{(\bm{x_j^{(t)}},\SDF_j(\bm{x_j^{(t)}})), 1 \leq t \leq T\}, \ 1 \leq j \leq N.
\label{eq:def_sub_datasets}
\end{equation}

\subsubsection{Learning the latent representation}
\label{subsubsection:learning_latent_representation}

In this paragraph we describe how the latent representation of the training shapes $\Bc_1,...,\Bc_N$ is learned by \textit{DeepSDF} model. 

\textit{DeepSDF} is an auto-decoder network \cite{auto_decoder2019}. In this architecture, there is no encoder network, \textit{i.e.} there is no parametric function $\Bc \mapsto \zb$ that directly encodes the shapes in the latent space.

\begin{figure}
    \centering
    \includegraphics[width=.6\linewidth]{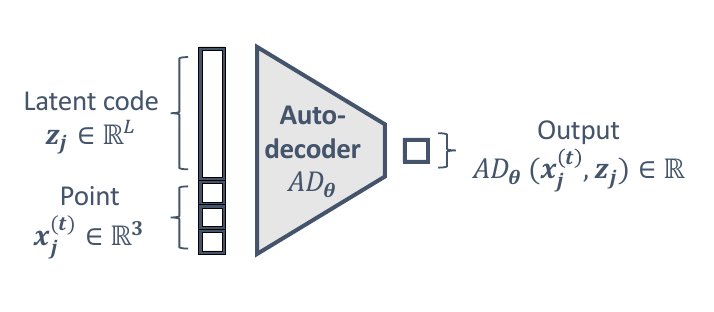}
    \caption{\textit{DeepSDF} model architecture. Each training input is the concatenation of a point $\bm{x_j^{(t)}}\in\R^3$ and a latent code $\bm{z_j}\in\R^L$ (the one referring to the envelope $\Bc_j$). The value of $L$ and the architecture of the network are hyperparameters of the model. The parameters optimized during the training step are both the latent codes $\bm{z_1},...,\bm{z_N}$ and the network's weights $\thetab$.}
    \label{fig:deepsdf_network_architecture}
\end{figure}

Let us denote the auto-decoder network $\AD_{\thetab}$. $\thetab$ refers to the weights of the network. The network architecture is depicted in Figure \ref{fig:deepsdf_network_architecture}. The input of the network is a concatenation of a latent code and three point coordinates. The output of the network is a scalar value.

Given a batch of points $\bm{x_j^{(t)}}\in\mathbb{R}^3$ and latent codes $\bm{z_j}\in\mathbb{R}^L$ ($1 \leq j \leq N$, $1 \leq t \leq T$), the network is trained in order to minimize a loss which is the absolute difference between its outputs and the true SDF values:
\begin{equation}
( \bm{\theta^{*}},\bm{z_1^{*}},...,\bm{z_L^{*}} ) \vcentcolon= \argmin_{\thetab,\bm{z_1},...,\bm{z_L}}\sum_{t,j}|\AD_{\thetab}(\bm{x_j^{(t)}}, \bm{z_j})-\SDF_k(\bm{x_j^{(t)}})| \label{eq:deepsdf_training_minimization}
\end{equation}

where $\bm{z_1},...,\bm{z_N} \in \R^L$ are the latent codes that refer to the shapes $\Bc_1,...,\Bc_N$. They are initialized with values sampled from a Standard Gaussian law in $\R^L$.

A regularized term is used to force the distribution of the latent codes to keep close to a Standard Gaussian distribution
\footnote{The main idea of the network optimization is summarized in Equation \ref{eq:deepsdf_training_minimization}. For the sake of simplicity, we consider that Equation \ref{eq:deepsdf_training_minimization} summarizes well the network optimization. For more details, we refer to Section 4.2 of \cite{park2019deepsdf}.}
.

Once the network has been trained, the latent codes $\bm{z_1},...,\bm{z_N}$ obtained constitute the latent representation of the training shapes $\Bc_1,...,\Bc_N$. In the context of regression or classification on the set of shapes, the latent codes are used as inputs to predict the labels associated with the shapes.

\subsubsection{Latent code inference for unseen shapes}
\label{subsubsection:inference}

Let us consider a new surface $\Bc_\mathrm{new} \notin \{\Bc_1,...,\Bc_N\}$. In more common approaches like autoencoder \cite{ballard1987modular}, we would be able to compute directly its latent representation using the encoder network as the latter maps any shape to a latent code.

In the auto-decoder framework, the latent code $\bm{z_{\mathrm{new}}}$ associated to $\Bc_\mathrm{new}$ is computed using inference \footnote{Here the term inference refers to Bayesian inference which is the underlying theory used here. The Bayesian derivation of the auto-decoder-based DeepSDF model is described in Section H of the supplementary material of \cite{park2019deepsdf}. In our work, we only describe the model practically. As explained right after, the latent code associated to the new shape is optimized using gradient descent.}
. Practically, a latent vector is randomly initialized and then optimized so that the output of the network gets as close as possible to the SDF values of the sampled points for the given new shape. Formally, points $\bm{x_{\mathrm{new}}^{(1)}},...,\bm{x_{\mathrm{new}}^{(T)}}$ are sampled close to the surface. The procedure used to sample these points is the same as the one used to build the datasets $\bm{X_1},...,\bm{X_N}$ (see Paragraph \ref{subsubsection:data_preparation}). The obtained dataset can be defined as follows:  \begin{equation}
\bm{X_\mathrm{new}} \vcentcolon= \{ (\bm{x_{\mathrm{new}}^{(t)}},\SDF_\mathrm{new}(\bm{x_\mathrm{new}^{(t)}})), 1 \leq t \leq T \}.
\label{eq:subdataset_for_new_shape}
\end{equation}

Afterwards, the previously trained auto-decoder network is used with fixed parameter values to perform a gradient descent in $\R^L$ as follows:
\begin{equation}
\bm{z_\mathrm{new}} \vcentcolon= \argmin\limits_{\bm{z}\in\R^L} \sum_t |\AD_{\thetab^{*}}(\bm{x_{\mathrm{new}}^{(t)}}, \bm{z})-\SDF_{\mathrm{new}}(\bm{x_{\mathrm{new}}^{(t)}})|. \label{eq:deepsdf_inference_minimization}
\end{equation}

\subsubsection{Shape generation}
\label{subsubsection:deepsdf_shape_generation}

Shape generation can be performed by picking some new point
\footnote{The way the point is picked depends on the application. For example, if the latent representation of the shapes is used as an input for a surrogate model of numerical simulations, one can pick a point in the latent space that is mapped to good simulation outputs.}
in the latent space $\R^L$. Given points randomly sampled in $\R^3$ in $\mathbb{R}^3$, one can get an estimate of the SDF at these points. Neverthelss, effective methods to compute the zero-level set of a given function $\R^3 \rightarrow \R$ are discussed in \cite{osher2004level, remelli2020meshsdf, liao2018deep}. Most of works in the field of computer vision use the Marching Cubes algorithm \cite{lorensen1987marchingcubes, lewiner2003efficient} on the Signed Distance Function. This algorithm is used to extract a 2D surface mesh from a 3D voxel grid. It can be conceptualized as a 3D generalization of isoline drawing on topographical maps. SDF values are incoded as a voxel grid and the algorithm looks for the voxels that cross the zero-level set of the SDF values.

	\subsection{Edge Detection}
        \label{subsection:intro_edge_detection}

A surface edge is typically characterized as a tangential discontinuity. In \cite{weber2010sharp}, Weber et al. describe edges as distinct features (such as peaks and valleys) where two planes meet, along with corners formed at the convergence of three or more planes. In \cite{hackel2016contour}, Hackel et al. refer to these as "wire-frame contours", defining edges as lines where there is an abrupt change in the orientation (normals) of the underlying surface.

In this paper, we propose a method for improving the training of Neural UDFs by focusing the network's effort on surface edges. Before training, and even before building the training dataset, all types of surface edges (crests, valleys, peaks,...) need to be detected. When the surface is encoded as a point cloud, it is conventional to compute local geometric descriptors at each point of the cloud \cite{himeur2021pcednet, demarsin2007detection, pauly2002efficient, weinmann2015curvature, hackel2016contour}.

For example, in \cite{pauly2002efficient}, Pauly et al. perform local covariance analysis at each point in a point cloud to compute several local geometric descriptors. They aim to detect surface edges. It allows them to sample more points around surface edges than in areas in which the surface is planar or quasi-planar. By doing so, they optimize the sampling of surfaces. This is called surface simplification. In \cite{weinmann2015curvature}, Weinmann et al. use these local geometric descriptors to perform classification and shape matching. 

Formally, consider a surface $\Bc$ numerically encoded by a point cloud $\Bb$. Let $k$ be a positive integer and let denote  $\bm{N_k}(\bm{x_0})\vcentcolon=(\bm{x_i})_{1 \leq i \leq k}$ the $k$ neighborhoods of a given point $\bm{x_0}$ belonging to the surface. That is,
\begin{equation}
\bm{N_k}(\bm{x_0}) \vcentcolon= \argmin_{\{\bm{y_1},...,\bm{y_k}\}\subset\Bb} \sum_{i=1}^k |\bm{y_i}-\bm{x_0}| \vcentcolon= \{\bm{x_1},...,\bm{x_k}\}.
\label{eq:k_neighbors}
\end{equation}

In this paper, $\bm{x_0}$ is referred to as the \textbf{centroid} of the \textbf{neighborhood} $\bm{N_k}(\bm{x_0})$. $k$ represents the scale at which we detect edges. It is an important parameter of the method. In \cite{himeur2021pcednet}, Himeur et al. propose to compute some local descriptors at several scales \textit{i.e.} for different values of $k$. In our work, $k$ is a fixed parameter and its value is empirically chosen. We discuss further on the the value of $k$ in Section \ref{subsection:results_edge_detection_parameters}.

Then, we can define the local covariance matrix
\begin{equation}
\bm{\mathcal{V}}=\frac{1}{k+1}\sum_{i=0}^k (\bm{x_i}-\bm{\overline{x}})(\bm{x_i}-\bm{\overline{x}})^T
\label{eq:covariance_matrix}
\end{equation}
where 
\begin{equation}
\bm{\overline{x}} \vcentcolon= \frac{1}{k+1}\sum_{i=0}^k \bm{x_i}
\label{eq:barycenter_of_neighbors}
\end{equation}
is the barycenter of $\bm{N_k}(\bm{x_0}) \cup \{\bm{x_0}\}$. The local covariance matrix (also called 3D structure tensor) is non negative-definite with three non-negative eigenvalues that correspond to an orthogonal system of eigenvectors. The three eigenvalues $\lambda_1 \geq \lambda_2 \geq \lambda_3$ of the local structure tensor give an idea of the local shape of the surface. In particular, \textbf{Pauly's descriptor} 
\footnote{In \cite{pauly2002efficient}, this feature is called surface variation. In this paper, it is both referred as surface variation and Pauly's descriptor} 
of the surface in $\bm{x_0}$ is given by the following formula:
\begin{equation}
\epsilon_{W}(\bm{x_0}) \vcentcolon= \frac{\lambda_3}{\lambda_1+\lambda_2+\lambda_3} 
\label{eq:weinmann_descriptor}
\end{equation} 
This local descriptor measures the surface variation with respect to the tangent plane of the local neighborhood $\bm{N_k}(\bm{x_0})$. If the surface is locally planar around $\bm{x_0}$, then $\lambda_3 \ll \lambda_1,\lambda_2$ and thus $\epsilon_W(\bm{x_0}) \simeq 0$. If $\bm{x_0}$ is close to a surface edge, then the local 3D structure tensor is more likely to be isotropic. In this case, $\lambda_1 \simeq \lambda_2 \simeq \lambda_3$ and $\epsilon_W(\bm{x_0}) \simeq \frac{1}{3}$.

As explained in \cite{weinmann2015curvature}, Pauly's descriptor, combined with other geometric indicators, yields great performance in point cloud classification tasks. But it turns out that in some cases, this method does not identify the surface edges. For example, if $\bm{x_0}$ is located close to a very acute crest or valley, then the local 3D structure is similar to the one of a planar area
\footnote{This limit of Pauly's descriptor is explained in more details in Appendix \ref{subsection:comparison_of_descriptors}.}
. Indeed, the average plane (carried by the two main axes of the eigen decomposition) can be orthogonal to the underlying surface, which invalidates the surface variation estimate made by Pauly's method. 

In this paper, we propose an edge detector that is also based on neighborhood selection, but that uses statistical tools. Indeed, our local descriptor is the $p$-value of a statistical test and not directly a geometric descriptor. It is described in details in Sections \ref{subsection:statistical_edge_detection} and \ref{subsection:detect_edges}.
	
	\subsection{Goodness-Of-Fit Tests For Sphericity}
        \label{subsection:central_symmetry_tests}

In this work, we propose a novel edge detection method. It is applied to a point cloud representing a surface. We compute a local descriptor at each point of the surface. This local descriptor is the $p$-value of a statistical test on the neighboring points after orthogonal projection on the average plane of the neighborhood (see Section \ref{subsection:detect_edges}). More precisely, the projected points are considered as \textit{i.i.d} random vectors in $\mathbb{R}^2$ (projection onto the average plane). In this Section, we give a review of the statistical tests commonly used to assess the sphericity of a distribution, based on the empirical cumulative distribution.

\subsubsection{Notations}

Let $\bm{X_1},...,\bm{X_k} \in \mathbb{R}^2$ be a set of \textit{i.i.d.} observations and $(r_1,\phi_1),...,(r_k,\phi_k)$ their polar coordinates \textit{w.r.t.} a given axis. We aim to test the following null hypothesis: 
\begin{equation}
H_0: \left\{
    \begin{array}{ll}
        H_{0,1}: \phi_1 \text{ is uniformly distributed on } [-\pi,\pi) \\
        H_{0,2}: r_1 \text{ and } \phi_1 \text{ are independent.}  
    \end{array}
\right.
\label{eq:null_hypothesis}
\end{equation}
Let us denote respectively $F_k^r$, $F_k^\phi$ and $F_k^{r,\phi}$ the empirical cumulative distribution functions (CDF) built on $(r_i)_{1 \leq i \leq k}$, $(\phi_i)_{1 \leq i \leq k}$ and $(r_i, \phi_i)_{1 \leq i \leq k}$ respectively. That is,

\begin{align}
\forall t \in \R_+, \  & F_n^r(t)\vcentcolon=\sum_{i=1}^k \mathbb{1}_{r_i\leq t}, \\
\forall \psi \in [-\pi,\pi), \ & F_k^\phi(\psi)\vcentcolon=\sum_{i=1}^k \mathbb{1}_{\phi_i\leq \psi},
\label{eq:empirical_cdf} \\
\forall (t,\psi)\in \R_+\times[-\pi,\pi), \ & F_k^{r,\phi}(t,\psi)\vcentcolon=\sum_{i=1}^k \mathbb{1}_{r_i\leq t  \ \mathrm{and} \  \phi_i\leq \psi}.
\end{align}

\subsubsection{Goodness-Of-Fit Tests For Uniformity of the Polar Angles}

The classical statistical methodology to validate or invalidate the null hypothesis $H_{0,1}$ (uniformity of the polar angles) consists to compute a discrepancy between the empirical distribution of the sample $(\phi_i)_{1 \leq i \leq k}$ and the uniform distribution. To begin with, let $F_0$ be the CDF of the uniform distribution. That is,
\begin{equation}
    \forall \psi \in [-\pi,\pi), \ F_0(\psi) = \frac{1}{2\pi}(\psi+\pi).
    \label{eq:th_cdf_uniform_minus_pi_pi}
\end{equation}
We will briefly discuss the classical goodness-of-fit test procedures.

This short review is based on the book \cite{nikitin1995asymptotic}. Following Y. Nikitin, we only focus on the test statistics that are based on the difference $F_k^\phi-F_0$.  In \cite{kolmogorov1933sulla}, Kolmogorov proposes the now classical statistic 
\begin{equation}
A_k\vcentcolon=\sup_{-\pi < \psi < +\pi}\left|F_k^\phi(\psi)-F_0(\psi)\right|
\label{eq:intro_ks_statistic}
\end{equation}

In \cite{darling1983}, Darling introduced the centered variant of the Kolmogorov statistics, which is usually called Watson-Darling statistic:
\begin{equation}
B_k\vcentcolon=\sup_{-\pi < \psi < +\pi}\left|F_k^\phi(\psi)-F_0(\psi)-\int_{-\pi}^{+\pi}(F_k^\phi(\rho)-F_0(\rho))dF_0(\rho)\right|
\label{eq:ks_statistic_invariant}
\end{equation}

Some other statistics are based on the $L^2$-norm of the difference $F_k^\phi-F_0$. For example the Cramér-Von-Mises-Smirnov statistic is defined as follow:
\begin{equation}
C_k\vcentcolon=\int_{-\infty}^{+\infty}(F_k^\phi(\psi)-F_0(\psi))^2dF_0(\psi)
\label{eq:cvm_statistics}
\end{equation}

These statistics tend to be more efficient than the Kolmogorov-Smirnov-based ones as they need smaller samples to converge. In the other hand, they involve integrating the squared difference between the empirical and theoretical CDFs, which can be computationally more intensive, especially for large sample sizes. Moreover, they are less sensitive to small deviations from the null hypothesis.

In \cite{andersondarling1952}, Anderson and Darling propose to generalize the previous statistic to any $L^\alpha$-norm  ($\alpha>0$) and to improve its properties by using a non-negative weight function $q$ as follows:
\begin{equation}
D_k^{\alpha,q}\vcentcolon=\int_{-\pi}^{+\pi}(F_k^\phi(\psi)-F_0(\psi))^\alpha q(F_0(\psi))dF_0(\psi).
\label{eq:anderson_darling_weighted}
\end{equation}

The incorporation of the weight function allows for more flexibility in handling data with varying importance or significance. This makes it suitable for situations where certain observations may be more influential than others in assessing the goodness-of-fit.

In terms of consistency and sensitivity to tail behavior, the best known weighted statistic is the Anderson-Darling statistic (see \cite{andersondarling1954}):
\begin{equation}
E_k\vcentcolon=\int_{-\pi}^{+\pi}\frac{(F_k^\phi(\psi)-F_0(\psi))^2}{F_0(\psi)(1-F_0(\psi))}dF_0(\psi).
\label{eq:anderson_darling_best}
\end{equation}

All the aforementioned test statistics can be compared in terms of Bahadur efficiency. The latter measures the rate at which the test statistic converges to its expected value under $H_0$. For the statistics described in this Section, the expected value under $H_0$ is zero. One can refer to \cite{bahadur1971some} for more details on Bahadur efficiency.

\subsubsection{Centered Cramer-Von-Mises Test On The Unit Circle} In \cite{smith1977} the author proposes a test statistics that estimates directly and globally the spherical symmetry of the underlying distribution. Let describe his procedure. First let introduce the following quantity:

\begin{equation}
X_k(r,\psi) \vcentcolon= F_k^{r,\psi}(r,\psi) - \frac{\psi}{2\pi}F_k^r(r)
\label{eq:def_X_k}
\end{equation}

In order that the test statistics does not depend on the chosen polar axis, he defines the following centered version:

\begin{equation}
Y_k(r,\psi) \vcentcolon= X_k(r,\psi) - \frac{1}{2\pi}\int_{0 \leq t \leq 2\pi} X_k(r,t) dt
\label{eq:def_Y_k}
\end{equation}

Under $H_0$, $Y_k(r,\psi)$ is close to $0$. This last observation motivates the following test statistic proposed in \cite{smith1977}:
\begin{equation}
U_k \vcentcolon= k \iint Y_k^2(r,\psi)dF_k^{r,\psi}(r,\psi).
\label{eq:smith_statistic}
\end{equation}

In our work, we use the Kolmogorov test statistic (see Equation \ref{eq:intro_ks_statistic}) applied to the angular coordinate of 2D points (see Section \ref{subsection:detect_edges}). We will proceed in order to avoid the drawback of a test statistics depending on the chosen origin on the circle (\textit{i.e.} the choice of polar axis $\phi=0$). Practically, instead of using the statistics proposed in Equation \ref{eq:ks_statistic_invariant}, we center the data using their Fréchet mean. The centering procedure is described in Section \ref{subsubsection:frechet_centering}.

The decision to validate or reject $H_0$ is based on the $p$-value of the test. It is defined as the probability that the test statistic $A_k$ (considered as a random variable) is greater or equal to its realization $a$, under $H_0$. That is:
\begin{equation}
\text{$p$-value} \vcentcolon= \P(A_k \geq a \ | \ H_0).
\label{eq:def_p_value}
\end{equation}

        \subsection{Distance between Point Clouds}
        \label{subsection:distance_between_point_clouds}

In this Section, we recall some classical distances between two point clouds. Let $\bm{A_1}\in\R^{n_1 \times D}, \bm{A_2}\in\R^{n_2 \times D}$ be two point clouds. Here, $D\in\{2,3\}$ is the dimension of the ambient space and $n_1,n_2$ are the number of points in $\bm{A_1}$ and $\bm{A_2}$. The three following distances are quite popular:

\begin{definition}[Hausdorff distance]
\begin{equation}
d_H(\bm{A_1},\bm{A_2}) \vcentcolon= \max( \max_{\bm{a}\in \bm{A_1}} d(\bm{a},\bm{A_2}), \max_{\bm{a}\in \bm{A_2}} d(\bm{a},\bm{A_1}) ).
\label{eq:hausdorff_distance}
\end{equation}
\label{def:hausdorff_distance}
\end{definition}

\begin{definition}[Chamfer distance]
\begin{equation}
d_{Ch}(\bm{A_1},\bm{A_2}) \vcentcolon= \frac{1}{n_1} \sum_{\bm{a}\in\bm{A_1}} d(\bm{a},\bm{A_2}) + \frac{1}{n_2} \sum_{\bm{a}\in\bm{A_2}} d(\bm{a},\bm{A_1}).
\label{eq:chamfer_distance}
\end{equation}
\label{def:chamfer_distance}
\end{definition}

\begin{definition}[Wasserstein distance]
\begin{equation}
d_{W}(\bm{A_1},\bm{A_2}) \vcentcolon= \min\limits_{\xi\in\text{Bij}} \sum_{\bm{a}\in\bm{A_1}} \|\bm{a}-\xi(\bm{a})\|_2.
\label{eq:wasserstein_distance}
\end{equation}
Here, $\text{Bij}$ denotes the set of all one to one applications from $\bm{A_1}$ onto $\bm{A_2}$.
\label{def:wasserstein_distance}
\end{definition}

In Geometric Deep Learning, it is a common practice to use Chamfer distance (Definition \ref{def:chamfer_distance}) or Wasserstein distance (Definition \ref{def:wasserstein_distance}) to train neural networks (by incorporating them into the loss calculation) or to evaluate them (by computing them after training) \cite{achlioptas2018learning, fan2017point, groueix2018papier, li2018so, park2019deepsdf}. These distances give an idea of the dissimilarity between two point clouds, averaged over all the underlying shapes. Here, we use the Hausdorff distance \cite{huttenlocher1993comparing} (Definition \ref{def:hausdorff_distance}) because it measures the local dissimilarity in the area where it is the highest. In the context of Neural UDF evaluation, it measures the highest local reconstruction error. In this sense, Hausdorff distance is considered more restrictive and is preferred from other distances.

    \section{Contributions}
    \label{section:contributions}

In our work, we propose a new statistical method to achieve edge detection on surfaces encoded as unstructured point clouds. We show that it can be used to improve the training procedure of 3D surface Neural UDFs.

In this Section, we formalize our main contributions. In Section \ref{subsection:statistical_edge_detection}, we describe our statistical edge detection method. In Section \ref{subsection:neural_udf_learning}, we discuss the task of UDF learning. Finally, the evaluation of the trained Neural UDFs is formalized in Section \ref{subsection:neural_udf_evaluation}.

	    \subsection{Statistical Edge Detection}
        \label{subsection:statistical_edge_detection}

In this Section, we discuss the problem of edge detection. Let us consider a watertight surface $\Bc \subset \R^3$ (see Definition \ref{def:watertight_surface}) encoded as a point cloud $\Bb \in \R^{n_s \times 3}$ containing $n_s$ points in $\R^3$. The points in the discrete set $\Bb$ are assumed to be uniformly sampled from the continuous surface $\Bc$.

Given a point $\bm{x_0} \in \Bc$ (this point is called {\bf centroid}), we wish to know if $\bm{x_0}$ is located on an edge of the surface. Let us extract from $\Bb$ the $k$ nearest neighbors of $\bm{x_0}$, as formalized in Equation \ref{eq:k_neighbors}: these points, denoted $\bm{x_1},...,\bm{x_k}$, constitute the \textbf{neighborhood} of $\bm{x_0}$ 

\begin{figure}
    \centering
    \includegraphics[width=\linewidth]{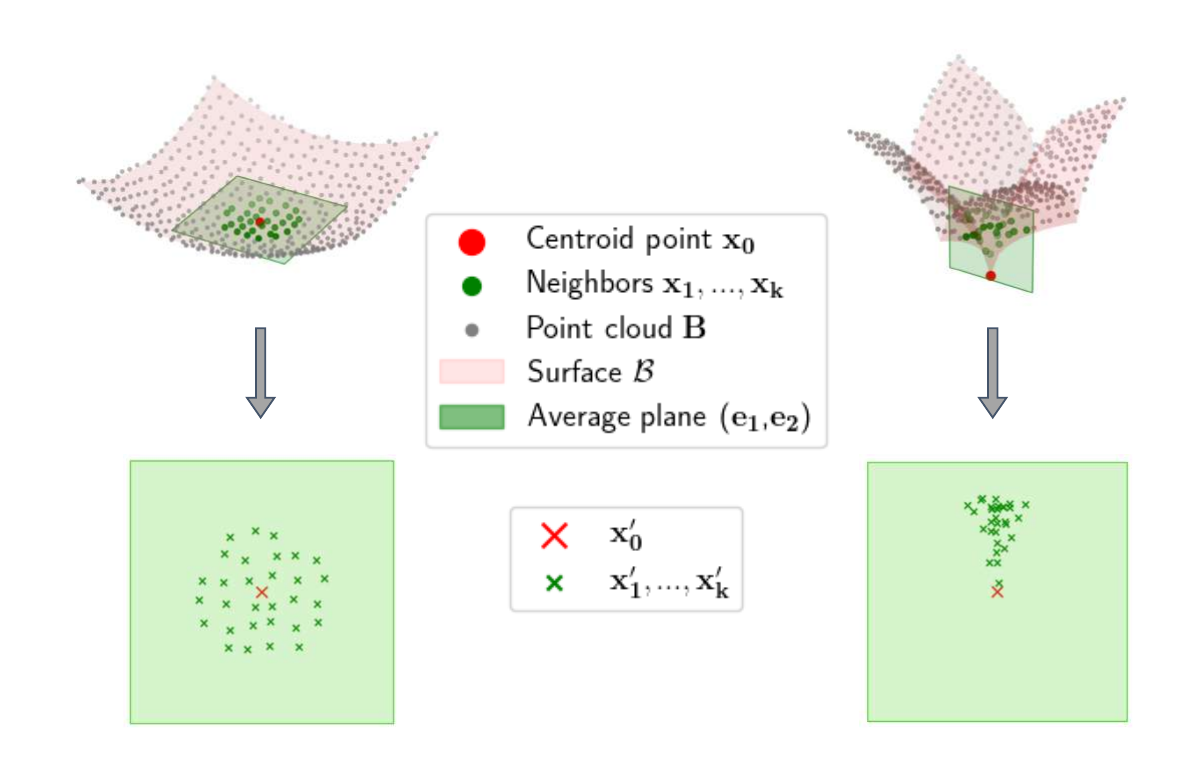}
    \caption{Projection onto the local average plane. On the left, a quasi-plane surface (\textit{situation 0}), and on the right, a pointed surface (\textit{situation 1}). On the top row, the surface $\Bc$ is in transparent red color. Points (in grey) are uniformly sampled from the surface. We aim to know if the surface is folded or pointed in $\bm{x_0}\in\Bc$. First, we compute its $k$ nearest neighbors $\bm{x_1},...,\bm{x_k}$ (green points). Then the average plane of the neighboring points is computed (in transparent green color). On the bottom row, the plots represent the projection of $\bm{x_0},...,\bm{x_k}$, denoted $\bm{x_0'},...\bm{x_k'}$. We observe that in \textit{situation 0}, the projection $\bm{x_0'}$ is in the middle of the neighbors' projections $\bm{x_1'},...\bm{x_k'}$. Whereas in \textit{situation 1}, it is completely off-centered.}
    \label{fig:edge_detection_scheme}
\end{figure}

The intuition that led to the edge detector described in this Section is inspired from the local surface variation descriptor proposed by Pauly et al. \cite{pauly2002efficient}. Our edge detector is based on an analysis of the projections of the points $\bm{x_0},...,\bm{x_k}$ on their \textbf{average plane} (defined in Section \ref{subsubsection:projection_onto_average_plane}). If the surface $\Bc$ is planar or quasi-planar (\textit{situation 0}) around $\bm{x_0}$, then the average plane is locally tangent to the surface. If the surface is very sharp (\textit{situation 1}) or folded around $\bm{x_0}$, then the average plane is not tangent to the surface locally (see Figure \ref{fig:edge_detection_scheme}). In order to differentiate between situations 0 and 1, we perform an orthogonal projection of the points $\bm{x_0},...,\bm{x_k}$ onto their average plane. This projection is formally described in Section \ref{subsubsection:projection_onto_average_plane}. The projected points are denoted $\bm{x_0'},...,\bm{x_k'}$. It can be visualized on Figure \ref{fig:edge_detection_scheme} that in \textit{situation 0}, $\bm{x_0'}$ lies in the middle of $\bm{x_1'},...,\bm{x_k'}$. Whereas in \textit{situation 1}, $\bm{x_0'}$ is off-centered with respect to the other projected points (see Figure \ref{fig:edge_detection_scheme}). 

The key point of our descriptor is to discriminate between these two situations using a statistical test of central symmetry on the projections of the points, with respect to the projection of $\bm{x_0}$ (that is considered as the potential center of symmetry under null hypothesis). We use the obtained $p$-value to discriminate between situations 0 and 1. This step is detailed in Section \ref{subsubsection:central_symmetry_test}.

The statistical test applied to the projections depends on reference frame for polar coordinates. In order to make it independent from this this reference frame, we center the angular coordinates with respect to their Fréchet mean. This is described in details in Section \ref{subsubsection:frechet_centering}.

\subsubsection{Projection of the neighboring points on the average plane}
\label{subsubsection:projection_onto_average_plane}

Given a set of points $\bm{x_0},...,\bm{x_k} \in \R^3$, we aim to compute their projections $\bm{x_0'},...,\bm{x_k'} \in \R^2$ on their average plane.

We define the average plane as the affine hyperplane of $\R^3$ containing the centroid point and orthogonal to the third eigenvector of the local covariance matrix. This matrix is computed on the points $\bm{x_0},...,\bm{x_k}$ according to Equation \ref{eq:covariance_matrix}. Roughly speaking, the average plane is the one in which the local point cloud extends the most (we refer to \cite{hotelling1933PCA} for more details on the average plane). Let us denote $\bm{e_1}$ and $\bm{e_2}$ the first two eigenvectors (with largest eigenvalues) of the local covariance matrix. The obtained 2D points can be defined as follows: 
\begin{equation}
\forall i \in \llbracket 0,k \rrbracket, \ \bm{x_i'} \vcentcolon= 
\begin{pmatrix}
\bm{x_i} \cdot \bm{e_1}
\\
\bm{x_i} \cdot \bm{e_2}
\end{pmatrix}
\label{eq:projection_onto_avg_plane}
\end{equation}




The next step consists in performing a central symmetry test on the set of points $\bm{x_0'},...,\bm{x_k'}$.

\subsubsection{Central symmetry test on the projected points}
\label{subsubsection:central_symmetry_test}

Given a set of points $\bm{x_0'},...,\bm{x_k'}$ on a hyperplane of $\R^3$, we aim to know if the points $\bm{x_1'},...,\bm{x_k'}$ are evenly distributed around $\bm{x_0'}$ (see Figure \ref{fig:edge_detection_scheme}).

Let us consider the centered vectors $(\bm{X_i})_{1 \leq i \leq k}$:
\begin{equation}
\forall i \in \llbracket 1,k \rrbracket, \ \bm{X_i}\vcentcolon=\bm{x_i'}-\bm{x_0'}.
\label{eq:centering_neighbors_wrt_ref_point}
\end{equation}

In our modelling we consider $\bm{X_1},...,\bm{X_k}$ as \textit{i.i.d} realizations of a 2D random vector. Our approach is to assess the central symmetry of this random vector around the origin. We rely on the method described in Section \ref{subsection:central_symmetry_tests}. We use Kolmogorov-Smirnov's test statistics $A_k$ defined in Equation \ref{eq:intro_ks_statistic} on the empirical CDF defined in Equation \ref{eq:empirical_cdf}, to validate or reject the null hypothesis $H_{0,1}$ (see (\ref{eq:null_hypothesis})). The $p$-value of the test is computed as in Equation \ref{eq:def_p_value}. Note that the polar angular coordinates are first centered with respect to their Fréchet mean. Kolmogorov-Smirnov's test is performed on the centered angles. Fréchet centering is described in details in Section \ref{subsubsection:frechet_centering}.

            \subsubsection{Kolmogorov-Smirnov's descriptor}
            \label{subsubsection:ks_descriptor}

If the $p$-value is higher than a given threshold $p_0$, we retain the null hypothesis, and consider that the angles are uniformly distributed. This supports the conclusion that the surface is locally quasi-planar. On the contrary, if the $p$-value is smaller than $p_0$, we conclude that the surface is pointed or folded (see Figure \ref{fig:edge_detection_scheme}).           
The computed $p$-value is thresholded using a hyperparameter $p_0 \in [0,1]$. This allows to define our statistical edge detector, also called \textbf{Kolmogorov-Smirnov's descriptor}:
\begin{equation}
\forall \bm{x} \in \Bb, \ \epsilon_{\text{KS}}(\bm{x}) \vcentcolon= 
\begin{cases}
1 \: \text{if \textit{$p$-value} $\leq p_0$.} \\
0 \: \text{if \textit{$p$-value} $> p_0$.}
\end{cases}
\label{eq:thresholded_descriptor}
\end{equation}

The threshold value $p_0$ is chosen empirically. We discuss the choice of $p_0$ in Section \ref{subsection:results_edge_detection_parameters}.

        \subsubsection{Fréchet Centering}
        \label{subsubsection:frechet_centering}

In our edge detection method, we calculate the $p$-value associated with a central symmetry test on the underlying distribution of the projected points in the average plane, seen as a $k$-sample $\bm{X_1},...,\bm{X_k} \in \R^2$. In order to do so, we use the polar coordinates of the observations. In particular, we perform a goodness-of-fit test on the empirical law of the polar angle (coordinate $\phi$) against the uniform law (as mentioned in the previous Section \ref{subsubsection:central_symmetry_test}, and formalized in Section \ref{subsection:central_symmetry_tests}).

The projected points $\bm{X_1},...,\bm{X_k}$ are initially written in the Cartesian reference system $(O,\bm{e_1},\bm{e_2})$ (see Equations \ref{eq:projection_onto_avg_plane} and \ref{eq:centering_neighbors_wrt_ref_point}). It is therefore natural to compute the polar coordinates $(r_1,\phi_1), ... ,(r_k,\phi_k)$ as follows:
\begin{equation}
\forall i \in \llbracket 1,k \rrbracket, \ 
	\begin{cases}
		r_k & \vcentcolon= \|X_k\|_2 \\
		\phi_k & \vcentcolon= \widehat{ \bm{e_1}, \bm{X_k} }
	\end{cases}
\label{eq:polar_coordinates}
\end{equation}
where $\widehat{\bm{e_1}, \bm{X}}$ is the oriented angle between $\bm{e_1}$ and $\bm{X}$, taken from the interval $[-\pi,\pi)$.

\begin{figure*}
    \centering
    \includegraphics[width=.99\linewidth]{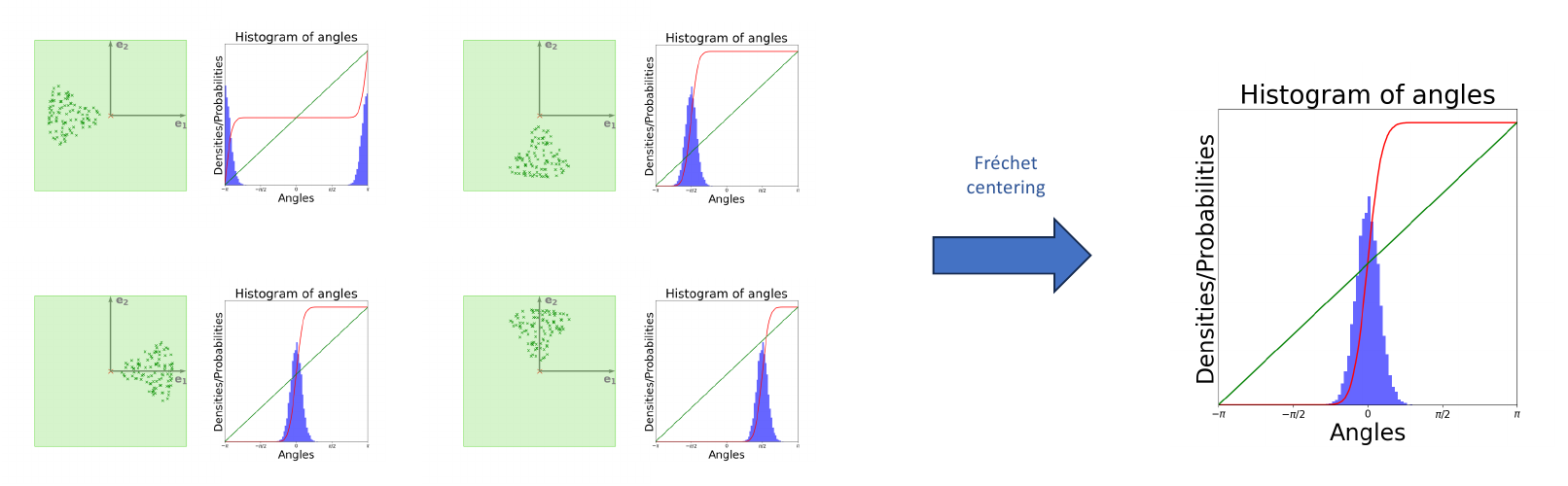}
    \caption{Fréchet centering of polar angles in the average hyperplane. The four green plots on the left correspond to the projections of the points of four point clouds onto their mean hyperplanes. In each situation, the projection of the centroid point is represented by a red cross in the center. The green crosses represent the projections of the neighboring points. In all four situations, the projection of the centroid point (red cross) is off-centered with respect to the projections of the neighboring points (green crosses). These are therefore four surface edges and the obtained $p$-values are expected to be roughly the same. However, the four point clouds are positioned differently with respect to the polar reference axis $\bm{e_1}$ (corresponding to $\phi=0$ in polar coordinates). The distribution of polar angles $\phi_1,...,\phi_k$ is therefore different in the four situations (see the four associated histograms). This results in different $p$-values for each situation. By centering the angles around their Fréchet mean, we obtain the same distribution centered at 0, as in the resulting histogram on the right. Thus, we obtain the same $p$-value in all four situations, which makes our method agnostic to any polar reference axis.}
    \label{fig:frechet_centering_scheme}
\end{figure*}

Thus, the polar angle (coordinate $\phi$) is defined with respect to a reference axis, and it turns out that the $p$-value of the aforementioned goodness-of-fit is highly dependent on this reference axis. Figure \ref{fig:frechet_centering_scheme} shows four different situations. They all correspond to a surface edge but the projected points are distributed in a different way on the average hyperplane (in the green rectangles). In the context of edge detection, we would expect to obtain roughly the same $p$-value for the four situations, but the reference axes $\bm{e_1}$ and $\bm{e_2}$ have different orientations, resulting in different angle distributions (on the histograms). 

To overcome this problem, we center the angles $\phi_1, ..., \phi_k$ with respect to their Fréchet mean \cite{frechet1948elements} before testing the central symmetry of the observations. For the sake of formalization, we define the angular distance 
\begin{equation}
\forall \psi_1,\psi_2 \in [-\pi,\pi), \  d_{\mathrm{ang}}(\psi_1,\psi_2) \vcentcolon= \min(|\psi_1-\psi_2|, 2\pi-|\psi_1-\psi_2|).
\label{eq:angular_distance}
\end{equation}

This distance can be conceptualized as the arclength distance between points along the unit circle. We use a generalization of the Euclidean mean, called Fréchet mean \cite{frechet1948elements}, on the angular distance:
\begin{equation}
\overline{\phi}_{F} \vcentcolon= \argmin_{\psi} \sum_{i=1}^n d_{\text{ang}}(\phi_i,\psi)^2.
\label{eq:frechet_mean}
\end{equation}

All angles are centered with respect to the Fréchet mean:
\begin{equation}
\forall i \in \llbracket 1,k \rrbracket, \ \phi_i' \vcentcolon= \phi_i-\overline{\phi}_{F}
\label{eq:frechet_centering}
\end{equation}

Visually, this amounts to using a reference axis that depends on the angles $\phi_1, ..., \phi_k$. More precisely, they are centered with respect to the average direction of the projected point cloud.

Note that the uniqueness of the Fréchet mean in not guaranteed for any set of observations. For example, if the angles $\phi_1,...\phi_k$ are distributed in a perfectly even way on $[-\pi,\pi)$, then their Fréchet mean is not well defined. In our work, we assume that this does not happen and that the Fréchet mean is always well defined. One can refer to \cite{charlier2013necessary} for a more detailed work on the uniqueness of the Frechet mean on the unit circle.

	\subsection{UDF Learning}
        \label{subsection:neural_udf_learning}

In this Section, we discuss the UDF learning task. It can be formulated as follows: given a watertight surface $\Bc \subset \R^3$ (Definition \ref{def:watertight_surface}), we aim to train a neural network to reproduce its UDF (Definition \ref{def:udf}). 

\begin{figure}
    \centering
    \includegraphics[width=.6\linewidth]{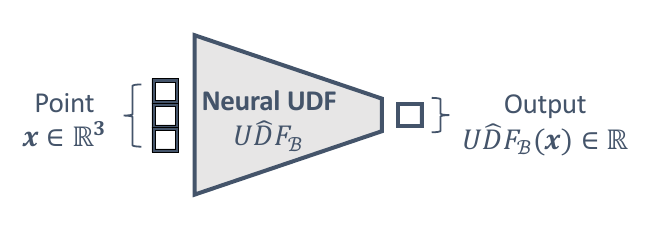}
    \caption{Neural UDF scheme for a watertight surface $\Bc$. The inputs of the network are tridimensional: they represents points in $\R^3$. The outputs are real values and the network is trained to fit the true UDF of the surface $\Bc$.}
    \label{fig:neural_udf_scheme}
\end{figure}

The function $\mathrm{UDF}_{\Bc}$ is an implicit representation of the surface $\Bc$ in the sense that it can not directly allow to draw the surface. Nevertheless, the {\it geometry} of the surface is contained in its UDF as its zero-level set:
\begin{equation}
\Bc = \mathrm{UDF}_{\Bc}^{-1}(\{0\}).
\label{eq:zero_levelset}
\end{equation}

Let us consider a fixed network architecture considered as a family of functions $\Fc$. The architecture is roughly described in Figure \ref{fig:neural_udf_scheme}. $\Fc$ is parameterized by the weights of the network. The \textbf{Neural UDF} training consists in the following optimization:
\begin{equation}
\widehat{\UDF}_{\Bc} \vcentcolon= \argmin\limits_{f\in\Fc} \sum_{\xb\in\Xb} (f(x)-\UDF_{\Bc}(\xb))^2.
\label{eq:neural_udf_training}
\end{equation}
where $\Bc$ and $\Xb \in \R^{n \times 3}$ is the training dataset. The procedure used to sample the training points $\Xb$ is discussed in details in Section \ref{subsection:sample_training_points}. $\widehat{\UDF}_{\Bc}$ is called the Neural UDF of the surface $\Bc$.

	\subsection{Neural UDF Evaluation}
    \label{subsection:neural_udf_evaluation}

In order to evaluate the Neural UDF of a given surface, we measure its ability to represent the surface. Consider a neural UDF $\widehat{\UDF}_{\Bc}$ that has been trained on a given surface $\Bc$. The evaluation task consists in quantifying the difference between the zero-level set of the Neural UDF and the true surface $\Bc$. In this Section, we explain how we estimate the Neural UDF's zero-level set and define the metric used to evaluate the neural UDF's accuracy.
    
For the sake of formalization, we define the zero-level set of the Neural UDF $\widehat{\UDF}_{\Bc}$ as follows:
\begin{equation}
\widehat{\Bc}\vcentcolon=\widehat{\UDF}_{\Bc}^{-1}(\{0\}).
\label{eq:estimated_zero_levelset}
\end{equation}

Effective methods to compute the zero-level set of a given function $\R^3 \rightarrow \R$ are discussed in Section \ref{subsubsection:deepsdf_shape_generation}. The mentioned methods require fine tuning to yield good results in a reasonable running time in our application. Moreover, they are commonly based on the Signed Distance Functions but here we only have access to the Unsigned Distance Functions.

In this work, we use another method: our idea is based on the fact that the Neural UDF is differentiable with respect to its inputs (spatial coordinates). A set $\bm{B_r}$ of $n_r$ points is sampled from the true surface $\Bc$ and is used as initial guess in the following optimization problem: 
\begin{equation}
\bm{\widehat{B}} \vcentcolon= \argmin\limits_{(\bm{b_1},...,\bm{b_{n_r}})\in\R^{n_r \times 3}} \sum_{i=1}^{n_r} |\widehat{\UDF}_{\Bc}(\bm{b_i})|.
\label{eq:gradient_descent_reconstruction}
\end{equation}

The optimization method is a gradient descent. The obtained set of points $\bm{\widehat{B}}$ is then compared to the initial set of points $\bm{B_r}$ sampled from the true surface. The difference between $\widehat{\Bc}$ and the true surface $\Bc$ is approximated using the Hausdorff distance between the point clouds $\bm{B_r}$ and $\bm{\widehat{B}}$. This metric is defined in Section \ref{subsection:distance_between_point_clouds}. Therefore, we define the \textbf{reconstruction error} of the Neural UDF as
\begin{equation}
\delta(\widehat{\UDF}_{\Bc}) \vcentcolon=  d_H(\bm{B_r},\bm{\widehat{B}}).
\label{eq:reconstruction_error}
\end{equation}

The reconstruction method we describe here has also some limitations. It highly depends on the initialization step \textit{i.e.} on the way $\bm{B_r}$ is built. Indeed, we initialize the points on the true surface $\Bc$. If the Neural UDF is accurate enough, its zero-level set can be assumed to be close to the true surface $\Bc$. This means that the estimated zero-level set $\bm{\widehat{B}}$ is potentially searched without exploring a large part of the ambient space $\R^3$. Furthermore, a large number of points are needed to cover the entire zero-level set of the Neural UDF.

    \section{Methodology}
    \label{section:methodology}

Neural UDF training and its evaluation consist of several steps. The steps that constitute our main contributions are discussed in the previous Section. In this Section, we present the step-by-step computational implementation of our method and and its evaluation.

Recall that,  given a surface $\Bc$, our aim is to train its Neural UDF. To measure the accuracy of the estimated Neural UDF, we need to compute the distance between the Neural UDF's zero-level set and the initial surface $\Bc$. Neural UDF training can be broken down into several sub-steps. First, we implement a function to detect pointed or folded areas (edges) on the surface (see Section \ref{subsection:detect_edges}). We then use this function to build a set of training points whose distribution is more concentrated around these edges (see Section \ref{subsection:sample_training_points}). 

Afterwards, we compute the true UDF of the surface at any point in space, using the mesh of $\Bc$ (see Section \ref{subsection:compute_true_udf}). Finally, we train a neural network to learn the true UDF. Once trained, this neural network is called a Neural UDF (see Section \ref{subsection:train_neural_udf}).

To measure the performances of our learning method, we use the trained Neural UDF to reconstruct the true surface and compare the obtained output with the true surface. This evaluation procedure is described in Section \ref{subsection:evaluate_method}. To perform this validation step, we generate points on the zero-level set of the Neural UDF. Next, we measure the distance between these points from the original surface. 

The whole pipeline for both training and evaluation of the Neural UDF is inherently random. Therefore, the results can vary from a run to another. In order to measure statistically the improvement due to our method, we repeat the experiments several times and compute the median value of the series of improvement measurements obtained. And we achieve this for several subsets of shapes from ShapeNet dataset \cite{chang2015shapenet}. This is explained in Section \ref{subsection:dealing_with_randomness}.

\begin{figure*}
    \centering
    \includegraphics[width=.99\linewidth]{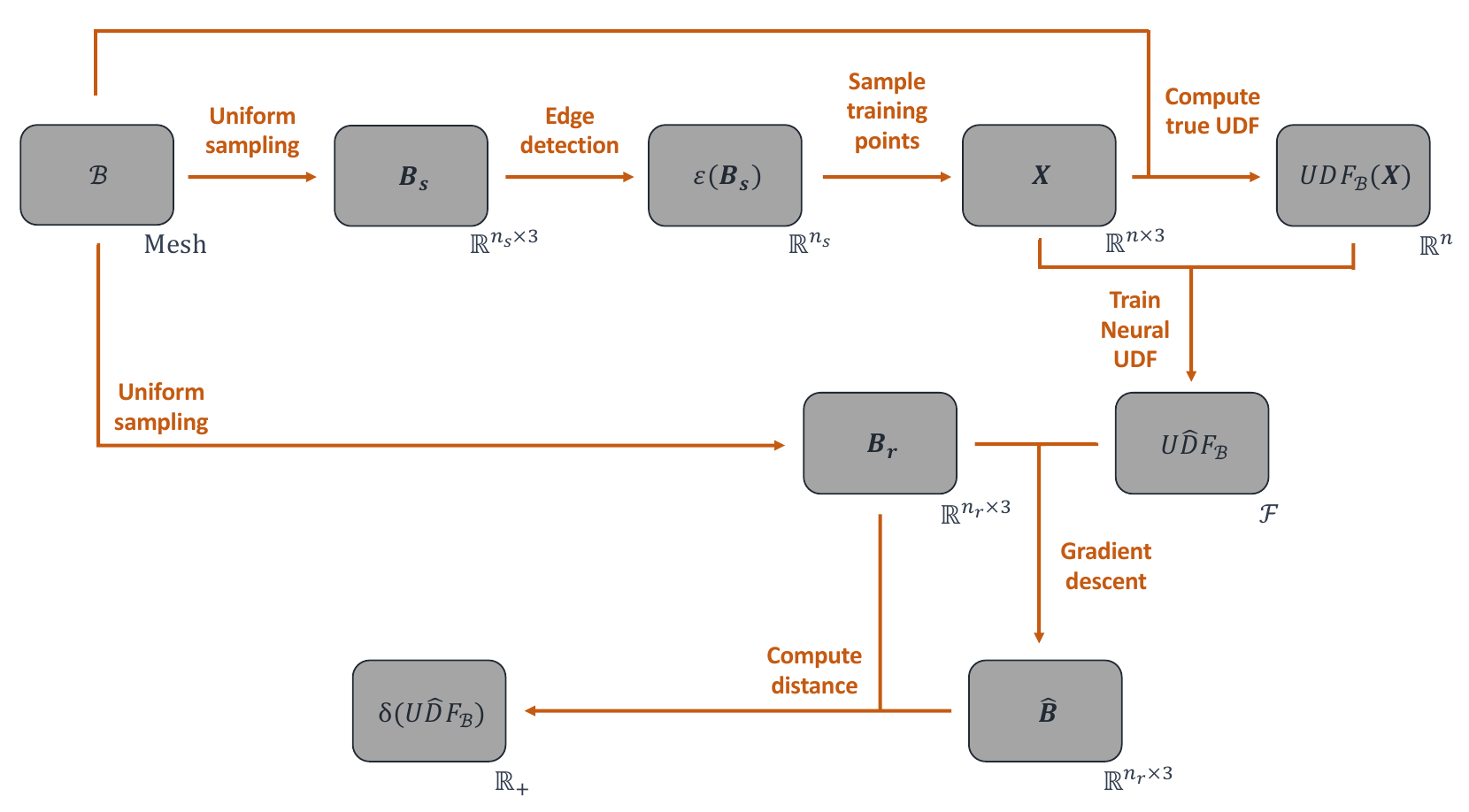}
    \caption{Neural UDF training and evaluation: methodology scheme. Given a 3D surface $\Bc$ encoded as a mesh, we aim to train its Neural UDF and evaluate its precision.}
    \label{fig:methodology_scheme}
\end{figure*}

	    \subsection{Detecting edges}
        \label{subsection:detect_edges}

In this Section, we describe the method used to detect edges on a surface and also explain how we choose the values of the involved parameters. Given a surface $\Bc$, the problem amounts in looking for the mapping:
\begin{equation}
\forall \bm{x}\in\Bc, \ \epsilon(\bm{x})\vcentcolon=
\begin{cases}
1 \: \text{if $\bm{x}$ is close to an edge of $\Bc$.} \\
0 \: \text{if $\Bc$ is locally planar or quasi-planar in $\bm{x}$.}
\end{cases}
\label{eq:edge_detection}
\end{equation}

First, we sample points uniformly from the surface. We denote $\bm{B_s}\in\R^{n_s \times 3}$ the point cloud thus generated. Given $\bm{x} \in \bm{B_s}$, we select the $k$ nearest neighbors of $\bm{x}$. The resulting points are then projected onto their average plane. A statistical test of central symmetry is performed on the resulting projections. The computed $p$-value is thresholded using a parameter $p_0 \in [0,1]$. This procedure is described in detail in Section \ref{subsection:statistical_edge_detection}. 

The three parameters $n_s$, $k$ and $p_0$ must be adjusted for each type of surface. For surfaces in ShapeNet dataset, we set $n_s=2000$, $k=40$ and $p_0=0.2$. The choice of these values is discussed in Section \ref{subsection:results_edge_detection_parameters}.
	
	    \subsection{Sampling the Training Points}
	    \label{subsection:sample_training_points}

In this Section, we describe the procedure used to sample the points used for Neural UDF training. We will also introduce and discuss the parameters used to configure this sampling procedure. We consider a watertight surface $\Bc$, and the same set of points $\bm{B_s}\in\R^{n_s \times 3}$ uniformly sampled on $\Bc$, as introduced in the previous section. It is assumed that the surface has been previously normalized in order to be a subset of the unit ball 
\begin{equation}
\Bc_{\|\cdot\|_2} = \{ \bm{x}\in\R^3, \ \|\bm{x}\| \leq 1 \}.
\end{equation}

Let denote the training dataset $\bm{X_0} \in \R^{n \times 3}$. It is an $n$-sized sample with probability distribution $\Lc(\R^3)$. Hereafter we define this probability distribution.

The training points need to be sampled everywhere in the unit ball $\Bc_{\|\cdot\|_2}$ and also on the surface $\Bc$. Hence, we build $\Lc(\R^3)$ as the mixture of two sub-distributions:
\begin{equation}
\Lc(\R^3) = (1-\nu) \Uc(\Bc_{\|\cdot\|_2}) + \nu \Lc(\Bc). 
\label{eq:tradeoff_ambient_surface}
\end{equation}
Here, $\Uc(\Bc_{\|\cdot\|_2})$ is the uniform distribution on the unit ball, $\Lc(\Bc)$ is a probability distribution on the surface $\Bc$ that is defined below (see Equation \ref{eq:distrib_on_surface}), and $\nu\in[0,1]$ is a parameter that quantifies how the training effort of the Neural UDF is concentrated on the surface $\Bc$. If $\nu=1$, the training points are all sampled on the surface. On the contrary, if $\nu=0$, the training points are all uniformly sampled from the unit ball $\Bc_{\|\cdot\|_2}$.

The points that are sampled from the surface can be either located near a surface edge or in a planar area. In order to improve the accuracy of the Neural UDF, we concentrate the training effort around surface edges, \textit{i.e.} around points $\bm{b}\in\bm{B_s}$ such that $\epsilon_{\text{KS}}(\bm{b})=1$. For the sake of formalization, let define the following sets:
\begin{equation}
\bm{B_{s,i}} \vcentcolon= \{\bm{b}\in\bm{B_s},\epsilon_{\text{KS}}(\bm{b})=i\}, i\in\{0,1\}.
\label{eq:define_B_s_0_B_s_1}
\end{equation}

In order to simplify the sampling procedure implementation, the points sampled from the surface are picked from the sets $\bm{B_{s,0}}$ and $\bm{B_{s,1}}$. Therefore, we can write
\begin{equation}
\Lc(\Bc) = (1-\nu_1(\Bc)) \Uc(\bm{B_{s,0}}) + \nu_1(\Bc) \Uc(\bm{B_{s,1}})
\label{eq:distrib_on_surface}
\end{equation}
where $\Uc(\bm{B_{s,0}})$ (\textit{resp.} $\Uc(\bm{B_{s,1}})$) is the uniform distribution on the point cloud $\bm{B_{s,0}}$ (\textit{resp.} $\bm{B_{s,1}}$) and $\nu_1(\Bc) \in [0,1]$ is a scalar that measures the proportion of the surface training points located close to surface edges.

In our experiments, we train Neural UDFs for several surfaces that can have different geometries. Some may have more edges than others. In other words, surfaces can have varying degrees of complexity. We formally define the complexity of the surface $\Bc$ as the value
\begin{equation}
\tau(\Bc) \vcentcolon= \frac{1}{n_s} \sum_{\bm{b}\in\bm{B_s}} \epsilon_{\text{KS}}(\bm{b}) = \frac{\mathrm{Card}(\bm{B_{s,1}})}{\mathrm{Card}(\bm{B_s})} 
\label{eq:surface_complexity_degree}
\end{equation}

To measure the influence of our surface edge detection method on the accuracy of Neural UDFs on a set of surfaces, we need to set a fixed oversampling parameter across all the surfaces. This oversampling parameter must locate the value $\nu_1(\Bc)$ between the degree of complexity $\tau(\Bc)$ (uniform sampling on the whole surface) and 1 (sampling on surface edges only). This amounts in setting a parameter $\xi \in [0,1]$ such that for any surface $\Bc$,
\begin{equation}
\nu_1(\Bc) = \xi + (1-\xi)\tau(\Bc).
\label{eq:def_nu_1}
\end{equation}

One can see that if $\xi=0$, then $\nu_1(\Bc)=\tau(\Bc)$, which means that the training points sampled on the surface are uniformly sampled on the whole surface (without oversampling the edges). On the contrary, if $\xi=1$, then $\nu_1(\Bc)=1$ and all the points sampled on the surface are sampled on surface edges.

In the dataset $\bm{X_0}$ constructed using the aforementioned sampling procedure, a large proportion of the points are sampled from the surface $\Bc$. Indeed, the laws $\Uc(\bm{B_{s,0}})$ and $\Uc(\bm{B_{s,1}})$ are distributions on the surface $\Bc$, since $\bm{B_{s,0}} \subset \Bc$ and $\bm{B_{s,1}} \subset \Bc$. Thus, the target output for these points is zero, as their UDF is equal to zero. To prevent the network from converging towards the trivial null solution, we perturb the points with a Gaussian noise. Hence the dataset rewrites as
\begin{equation}
\bm{X} \vcentcolon= \bm{X_0} + \bm{e}
\label{eq:definition_of_X}
\end{equation}
where $\bm{e} \in \R^{n \times 3}$ is a $(n,3)$-sample of the distribution $\Nc(0,0.025^2)$. This Gaussian perturbation and the value of the standard deviation $0.025^2$ has been suggested in the supplementary material of \cite{park2019deepsdf}. 

        \subsection{Computing the True UDF}
        \label{subsection:compute_true_udf}

In this Section, we describe the method used to calculate the true Unsigned Distance Function of a watertight surface $\Bc$ at training points $\Xb$. Formally, we seek to compute 
\begin{equation}
\forall \xb \in \Xb, \ \mathrm{UDF}_{\Bc}(\xb) = \min_{\bb \in \Bc} \| \xb-\bb \|
\label{eq:compute_true_UDF}
\end{equation}

To do this, we use the function \texttt{find\_closest\_cell} from framework \texttt{PyVista} \cite{sullivan2019pyvista}. For a query point $\xb\in\R^3$ and a surface $\Bc$ encoded as a triangular surface mesh, this function returns the point $\bm{b_x}\in\Bc$ closest to $\xb$. $\bm{b_x}$ can be seen as the orthogonal projection of $\bm{b}$ in $\Bc$. Once $\bm{b_x}$ has been found, we compute the distance $\|\xb-\bm{b_x}\|$ to find $\mathrm{UDF}_{\Bc}(\xb)$.

The function \texttt{find\_closest\_cell} used here is not optimally implemented. This is not a problem for us, as we calculate the value of $\UDF_{\Bc}$ for sets of points of size $n\simeq1000$. To reduce the time needed to calculate $\UDF_{\Bc}$, we can sample points uniformly over $\Bc$ and index these points using a $k$-dimensional tree \cite{bentley1975multidimensional}.

        \subsection{Training the Neural UDF}
        \label{subsection:train_neural_udf}

\begin{figure}
    \centering    \includegraphics[width=.9\linewidth]{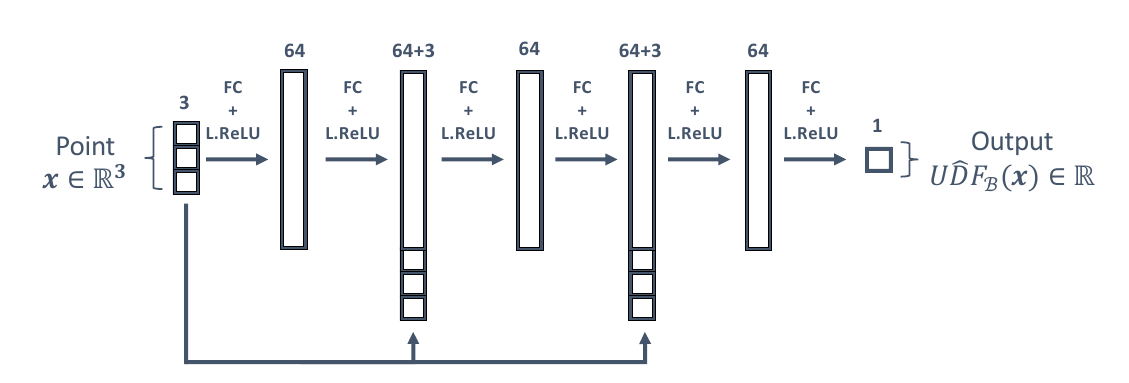}
    \caption{Neural UDF architecture. The network is composed of three blocks. Each of the three is composed of two fully-connected layers (\texttt{Leaky ReLU} activation functions \cite{maas2013rectifier}). The blocks are separated with two skip connections \cite{he2016deep} to avoid gradient vanishing issues.}
    \label{fig:neural_udf_architecture}
\end{figure}

In our work, Neural UDFs are Multi Layer Perceptron neural networks with 3 input features (input point coordinates) and 1 output feature (input point UDF). Their architecture is described in Figure \ref{fig:neural_udf_architecture}.

During network training, weights are optimized with the \textit{Adam} gradient descent \cite{kingma2014adam}. The training inputs are the points in the dataset $\Xb$ (see Equation \ref{eq:definition_of_X}) and the target outputs are their true UDF values (see Equation \ref{eq:compute_true_UDF}). The minimization is formally written in Equation \ref{eq:neural_udf_training}.
 
	    \subsection{Evaluating the Reconstruction Capacity of a Neural UDF}
        \label{subsection:evaluate_method}

Consider a surface $\Bc$ and its trained neural UDF $\widehat{\UDF}_{\Bc}$. In this Section, we quickly review the computational implementation of the evaluation of the neural UDF's accuracy.  
The evaluation procedure and the resulting metric are described in detail in Section \ref{subsection:neural_udf_evaluation}.

First, we construct a point cloud representing the surface $\Bc$. To do this, we sample points uniformly on $\Bc$ and gather them in the set $\bm{B_r} \in \R^{n_r \times 3}$, with $n_r=2000$. These points are used in the initialization step to reconstruct the zero-level set of the Neural UDF $\widehat{\UDF}_{\Bc}$, according to Equation \ref{eq:gradient_descent_reconstruction}. We denote $\bm{\widehat{B}}$ the reconstructed points. 

The ability of the Neural UDF to reconstruct the true surface $\Bc$ is given by the distance between $\bm{B_r}$ and $\bm{\widehat{B}}$ (see Equation \ref{eq:reconstruction_error}).

        \subsection{Dealing with Randomness}
        \label{subsection:dealing_with_randomness}

Neural UDF training is an inherently random procedure. First, to detect surface edges, points are randomly generated on the surface (see Section \ref{subsection:detect_edges}). Second, the Neural UDF training dataset is an $n$-sample of a mixture of probability distributions (see Section \ref{subsection:sample_training_points}). The initialization of the Neural UDF's weights is also randomized. 

Furthermore, Adam gradient descent \cite{kingma2014adam} and mini-batch learning \cite{bertsekas1996incremental} are used for both the Neural UDF's weights optimization (in the training procedure) and the surface reconstruction (in the evaluation procedure). In this optimization method, each iteration is randomized.

To take into account the randomness in the modelling, the experiments are ran on a set surfaces. And for each surface, Neural UDF training and evaluation are repeated several times and we calculate the median of the reconstruction errors. Formally, consider a set of surfaces $\Bc_1,...,\Bc_N$, an integer $n$ representing the number of training points and $\xi \in [0,1]$ the surface edges oversampling parameter, as defined in Section \ref{subsection:sample_training_points}. For each of the surfaces, we train 5 Neural UDFs with different random seeds. The 5 computed values of the metrics are summarized in their median.

We denote $\left\{\widehat{\UDF}_{\Bc_j}^{l}(n,\xi)\right\}_{
\substack{
1\leq j \leq N \\
1 \leq l \leq 5
}}$ the Neural UDFs obtained, and their reconstruction errors $\left\{\delta_{\Bc_j}^{l}(n,\xi)\right\}_{
\substack{
1\leq j \leq N \\
1 \leq l \leq 5
}}$ (as defined in Equation \ref{eq:reconstruction_error}). 

As we aim to measure the influence of surface edge oversampling on the reconstruction error, we repeat the same operation with $\xi=0$. Thus, we obtain a set of Neural UDFs $\left\{\widehat{\UDF}_{\Bc_j}^{l}(n,0)\right\}_{
\substack{
1\leq j \leq N \\
1 \leq l \leq 5
}}$ and their reconstruction errors $\left\{\delta_{\Bc_j}^{l}(n,0)\right\}_{
\substack{
1\leq j \leq N \\
1 \leq l \leq 5
}}$. For each of the Neural UDFs and for a fixed number of training points $n$, we compute the \textbf{relative improvement} due to $\xi$ :
\begin{equation}
\forall j \in \llbracket 1,N \rrbracket, \ \Ic(\Bc_j,n,\xi) \vcentcolon=  1 - \frac{\delta^m_{\Bc_j}(n,\xi)}{\delta^{m}_{\Bc_j}(n,0)}
\label{eq:precision_improvement}
\end{equation}
where $\delta^m_{\Bc_j}(n,\xi)$ is the median reconstruction error of the five Neural UDFs of shape $\Bc_j$ trained with parameter values $n$ and $\xi$, that is:
\begin{equation}
\delta^m_{\Bc_j}(n,\xi) \vcentcolon= \mathrm{median} \left(\left\{\delta_{\Bc_j}^{l}(n,\xi)\right\}_{1 \leq l \leq 5}\right)
\label{eq:runwise_median_xi}
\end{equation}
and naturally,
\begin{equation}
\delta^m_{\Bc_j}(n,0) \vcentcolon= \mathrm{median} \left(\left\{\delta_{\Bc_j}^{l}(n,0)\right\}_{1 \leq l \leq 5}\right).
\label{eq:runwise_median_0}
\end{equation}

\section{Results}

In this Section, we present the results of our numerical experiments. In Section \ref{subsection:results_edge_detection_parameters}, we analyze the influence of the scaling parameter $k$ (see Section \ref{subsection:statistical_edge_detection}) and the thresholding parameter $p_0$  on edge detection accuracy (see Section \ref{subsection:detect_edges}). In Section \ref{subsection:results_edge_detection} we show that our descriptor detects surface edges more accurately than Pauly's descriptor \cite{pauly2002efficient} (see Section \ref{subsection:intro_edge_detection} for a detailed explanation of this descriptor). In Section \ref{subsection:precision_on_edges}, we show that oversampling surface edges improves the accuracy of Neural UDFs locally around surface edges. In Section \ref{subsection:results_reconstruction_capacity}, we show that improving the accuracy of Neural UDFs around surface edges improves the global reconstruction capability of Neural UDFs.

    \subsection{Discussion on the Parameters for Edge Detection}
    \label{subsection:results_edge_detection_parameters}

Our edge detector is described in details in Section \ref{subsection:statistical_edge_detection}. It depends on the values of three parameters:
\begin{itemize}
    \item $n_s$ is the number of points sampled on the surface.
    \item $k$ is the number of nearest neighbors in neighborhood selection. It can be considered as a scale parameter.
    \item $p_0$ is the decision threshold for the $p$-value
\end{itemize}

Of course, the choice of $n_s$ should depend on the complexity of the considered shapes and the available computational resources. Indeed, for simple shapes containing few folded or pointed areas, a few points seem enough to depict the shape. For more complex surfaces with numerous edges, we need to sample more surface points. However, our edge detection method requires to perform a test and compute a $p$-value for each sampled point. Therefore, it can be computationally costly to compute our descriptor for a large number of surface points. In our work, and in particular on ShapeNet dataset, we empirically observe that $n_s=2000$ points are generally sufficient to describe accurately the surfaces and require a reasonable computation time.

\begin{figure}
  \centering
  
    \includegraphics[width=.7\linewidth]{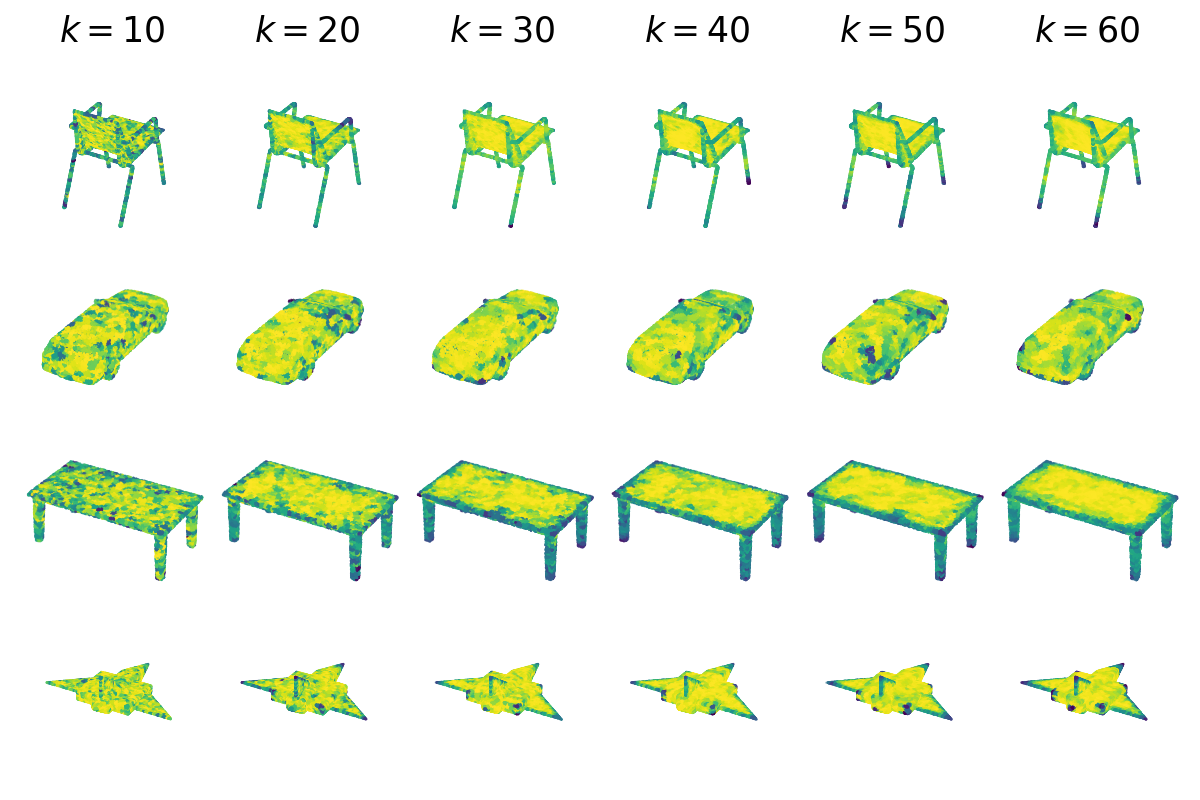}

    \caption{Setting the scale parameter $k$. Kolmogorov-Smirnov's descriptor on a few shapes on the dataset ShapeNet (\texttt{chair}, \texttt{car}, \texttt{table} and \texttt{airplane}) for several values of $k$. The local descriptor that is depicted here is the log-$p$-value of the local central symmetry test (see Equation \ref{eq:def_p_value}). It can be seen on the plots that edges are detected for high values of $k$ (higher than 30). All the edge detectors are computed with $n_s=2000$ points. The log-$p$-value is not thresholded here.}
    
  \label{fig:hyperparameter_k}
  
\end{figure}

The parameter $k$ corresponds to the scale at which the surface edges are detected. The choice of the scale is important because a point cloud obtained through an acquisition process consists of points that sample real-world objects, where edges can be rounded or damaged (we refer to \textit{rounding} and \textit{damage} as defined in \cite{himeur2021pcednet}). The scale at which edges can be detected depends of their degree of rounding and damage. For example, in a building, two façades may be joined by a smoothly curved surface, perceived as an edge at the building scale, which might not be readily detectable at finer scales, such as the scale of individual bricks \cite{himeur2021pcednet}. In our work, the scale parameter $k$ is the same for all the shapes of a dataset. We choose it empirically by comparing visually its the best value on a few examples. In Figure \ref{fig:hyperparameter_k}, one can observe that edges are detected for values of $k$ higher than 30. In all our experiments, we set $k=40$.

\begin{figure}
  \centering
  
  \includegraphics[width=.7\linewidth]{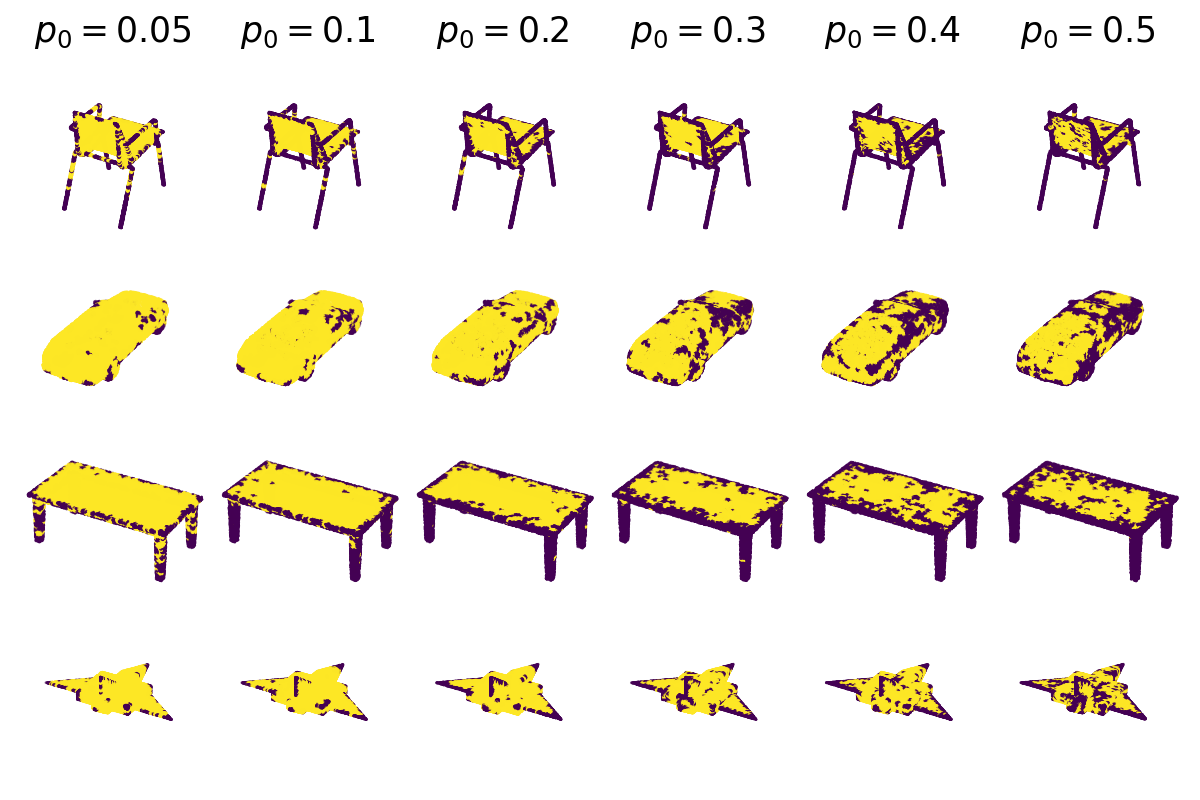}

    \caption{Setting the thresholding parameter $p_0$. Kolmogorov-Smirnov's descriptor on a few shapes on the dataset ShapeNet (\texttt{chair}, \texttt{car}, \texttt{table} and \texttt{airplane}) for several values of $p_0$. The local descriptor that is depicted here is the thresholded $p$-value $\epsilon_{n_s,k,p_0}$ (see Equation \ref{eq:thresholded_descriptor}). It can be seen on the plots that edges are detected when $0.4 \leq p_0 \leq 0.6$. All the edge detectors are computed with $n_s=2000$ and $k=40$.}
    
  \label{fig:hyperparameter_p_0}
  
\end{figure}

As highlighted in Equation \ref{eq:thresholded_descriptor}, Kolmogorov-Smirnov's descriptor is strongly dependent on a decision threshold that we call $p_0$. It corresponds to the minimal $p$-value for which we retain the null hypothesis \textit{i.e.} consider that the surface is locally planar or quasi-planar. The value of this threshold parameter is set empirically by comparing the plots of a few shapes. For the sake of automation, it remains the same for all the shapes of the dataset. It turns out that for categories \texttt{chair}, \texttt{car}, \texttt{table} and \texttt{airplane} of ShapeNet dataset, edges are well detected for $0.05 \leq p_0 \leq 0.3$.

\subsection{Edge Detection On ShapeNet Data}
\label{subsection:results_edge_detection}

In order to evaluate the accuracy of our edge detector, we compare it with the reference descriptor proposed in \cite{pauly2002efficient} (see Section \ref{subsection:intro_edge_detection}). The plots show that our edge detector is more accurate on chairs, cars, tables and airplanes in the ShapeNet dataset (see Figure \ref{fig:hyperparameter_p_0}). This difference is mainly due to two limitations of Pauly's descriptor. First, for very sharp or locally folded surfaces, Pauly's descriptor approaches zero because the local neighborhood of points can be almost as flat as for quasi-planar areas (this issue is illustrated in Figures \ref{subfig:results_toy_cones} and \ref{subfig:results_toy_folds}). Secondly, thin plates are often detected as edges by Pauly's descriptor because the neighborhood can contain both faces of the plate (see Figure \ref{subfig:results_toy_plates}). These two limitations are explained with more details and illustrated on toy examples in Appendix \ref{sup_mat:compare_ks_weinmann}.

\begin{figure}
  \centering
  
  \includegraphics[width=.99\linewidth]{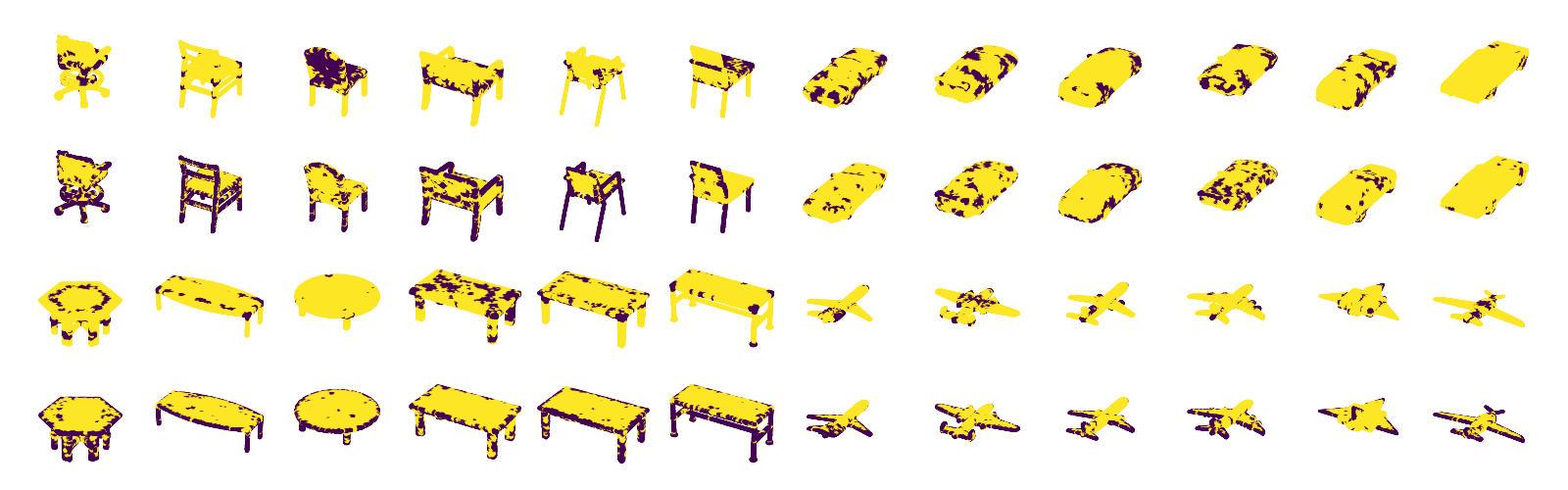}

  \caption{Edge detection using Pauly's descriptor (rows 1 and 3) \textit{vs.} Kolmogorov-Smirnov's descriptor (rows 2 and 4). Surfaces are sampled with $n_s=2000$ points. Each point on the surface is associated with $k=40$ neighboring points. Points with a $p$-value lower than $p_0=0.4$ are considered as close to an edge. On the plots, the latter are colored in purple. The undetected areas are colored in yellow. }

  \label{fig:edge_detection_plots_shapenet}
  
\end{figure}

\subsection{Application to UDF Learning}
\label{subsection:results_application_to_neural_udf_learning}
    
\subsubsection{Precision on Edges}
\label{subsection:precision_on_edges}

In this Section, we show that our method improves the accuracy of Neural UDFs locally around the surface edges. 

Let us consider a surface $\Bc$ and a Neural UDF $\widehat{\UDF}_{\Bc}$ trained with parameters $n$ and $\xi$ defined in Section \ref{subsection:sample_training_points}. Points are uniformly sampled on the surface and Kolmogorov-Smirnov's descriptor is computed for each of them. Doing this, we are repeating the same procedure than the one described in Section \ref{subsection:sample_training_points}. Among the set of points uniformly sampled, we call $\bm{B_{s,0}}$ (\textit{resp.} $\bm{B_{s,1}}$) the set of points that are not detected (\textit{resp.} detected) as edges (see Equation \ref{eq:define_B_s_0_B_s_1}). 

In order to measure the accuracy of Neural UDFs around these surface edges, we compute the average magnitude of the Neural UDF's output over the edges:

\begin{equation}
\overline{|\widehat{\UDF_{\Bc}}(\bm{B_{s,1}})|} \vcentcolon= \frac{1}{\mathrm{Card}(\bm{B_{s,1}})} \sum_{\bm{b} \in \bm{B_{s,1}}} |\widehat{\UDF_{\Bc}}(\bm{b})|
\label{eq:avg_neural_udf_on_edges}
\end{equation}

This metric can be considered as the average error of the Neural UDF on the edges, because the target output for these points is zero as they lie on the surface $\Bc$. 

\begin{figure}[htbp]
    \centering
    \includegraphics[width=.7\linewidth]{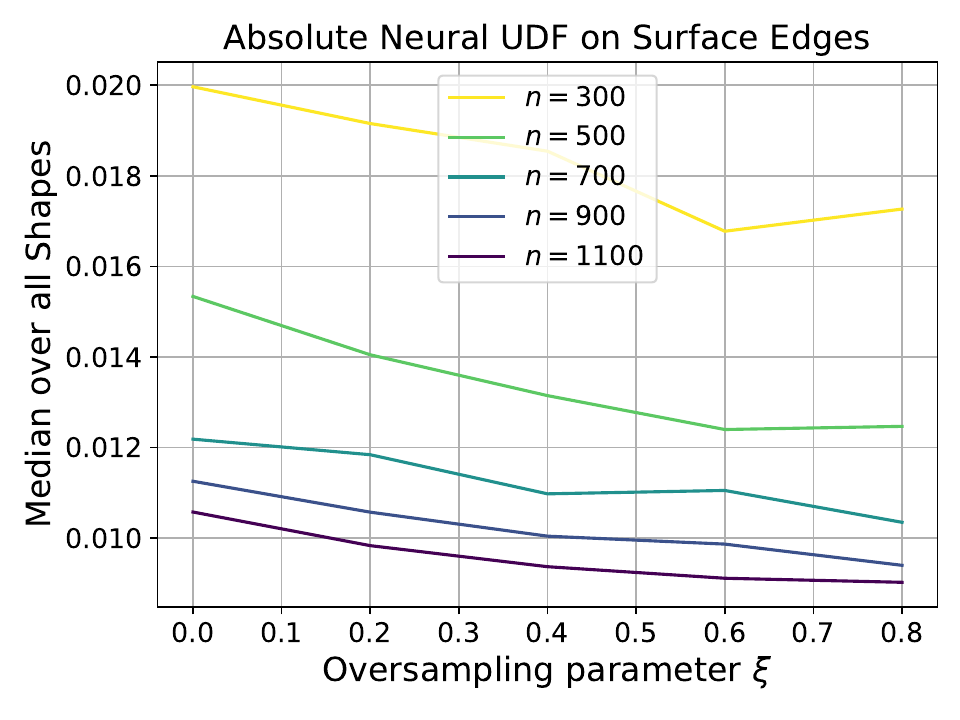}
    \caption{Evolution of the average Neural UDF error near edges (see Equation \ref{eq:avg_neural_udf_on_edges}) with respect to the oversampling parameter $\xi$, for several values of the number of training points $n$. The plotted metric corresponds to the median of the values computed for each of the Neural UDFs, over $N=25$ chairs from ShapeNet dataset.}
    \label{fig:neural_udf_on_edges}
\end{figure}

Figure \ref{fig:neural_udf_on_edges} illustrates that the Neural UDF error on surface edges decreases as $\xi$ increases, i.e. as surface edges are oversampled. This shows that oversampling improves the Neural UDF locally. On the other hand, we observe that this local improvement in accuracy decreases as $n$ increases. This is because, when a very large number of training points are sampled, the folds and peaks are already well sampled and oversampling them has less impact.

        \subsubsection{Global precision}
        \label{subsection:results_reconstruction_capacity}

In this Section, we show that our edge detection method can be used to improve Neural UDF training. The main idea is to sample more training points around the surface edges, as formalized in Section \ref{subsection:sample_training_points}. As a result, the accuracy of the network is locally better around these areas (see Section \ref{subsection:precision_on_edges}.

Nevertheless, in the frame of UDF learning, the aim is to accurately reconstruct the entire surface. So that,  the distance between initial and reconstructed surfaces is an interesting metric. This reconstruction error is defined in Section \ref{subsection:evaluate_method} (see Equation \ref{eq:reconstruction_error}).

As UDF learning and its evaluation are subject to a high degree of randomness, the precision metrics may vary from one run to another. We show that the metrics are statistically improved by repeating the experiments several times and keeping the median of the computed metrics. This approach is described in detail in Section \ref{subsection:dealing_with_randomness}.

We apply our method to ShapeNet dataset, for 4 categories (chair, car, table and airplane). For each category, we train Neural UDFs for $N=25$ different shapes, randomly taken from the dataset. For each shape, we compute the accuracy improvement due to the use of Edge Detection to build the training dataset (see Equation \ref{eq:precision_improvement}). 

\begin{figure}[htbp]
    \centering
    \begin{subfigure}{0.49\linewidth}
        \centering
        \includegraphics[width=\linewidth]{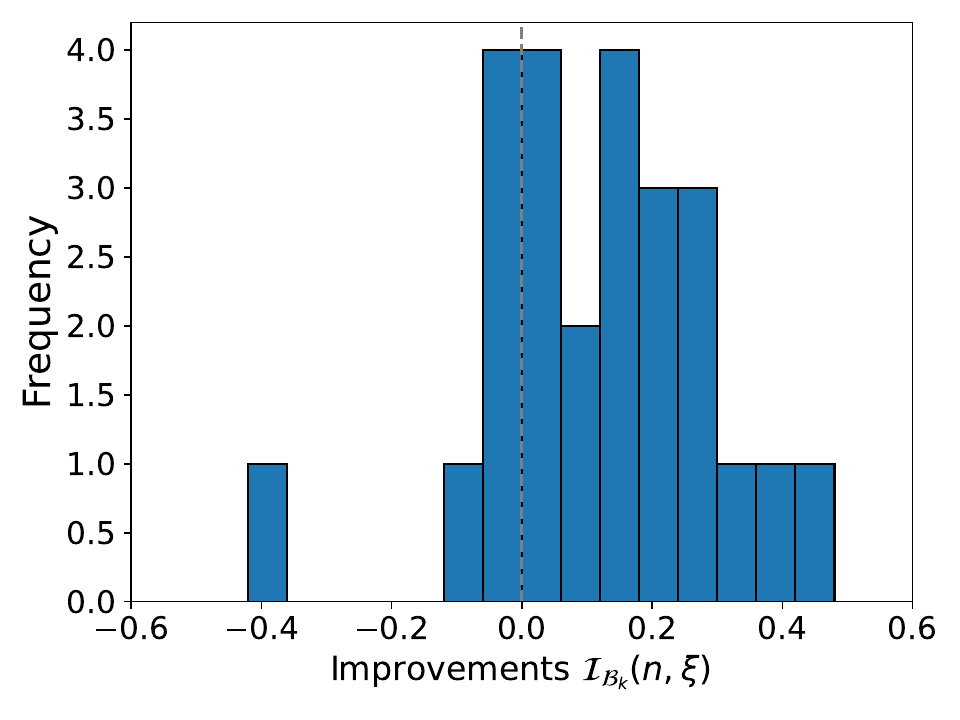}
        \caption{Chairs}
        
    \end{subfigure}
    \hfill
    \begin{subfigure}{0.49\linewidth}
        \centering
        \includegraphics[width=\linewidth]{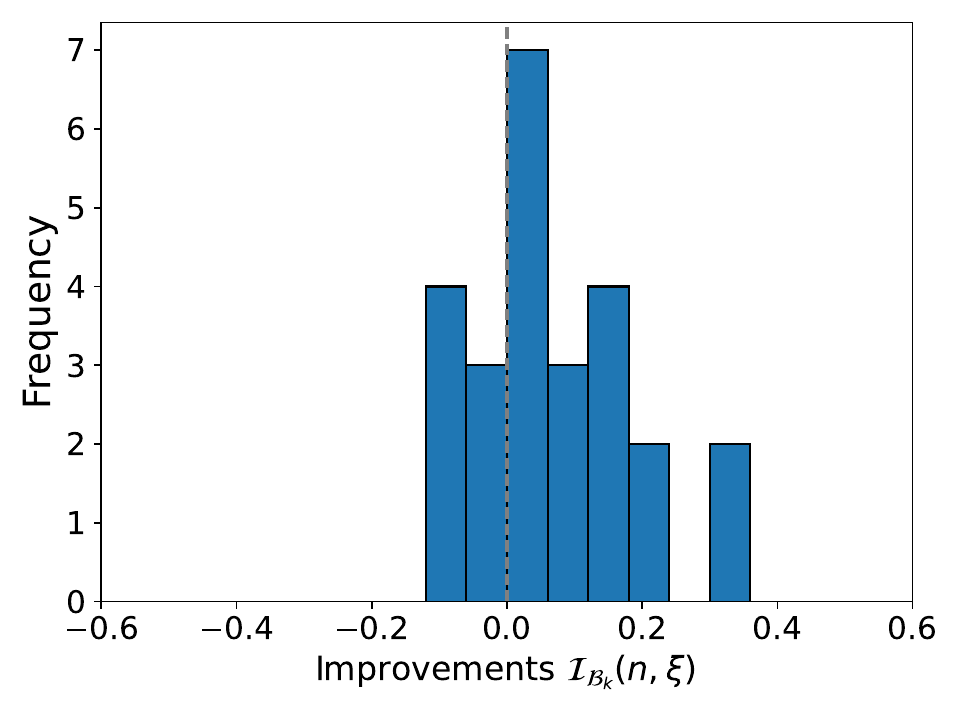}
        \caption{Cars}
        
    \end{subfigure}
    \hfill
    \begin{subfigure}{0.49\linewidth}
        \centering
        \includegraphics[width=\linewidth]{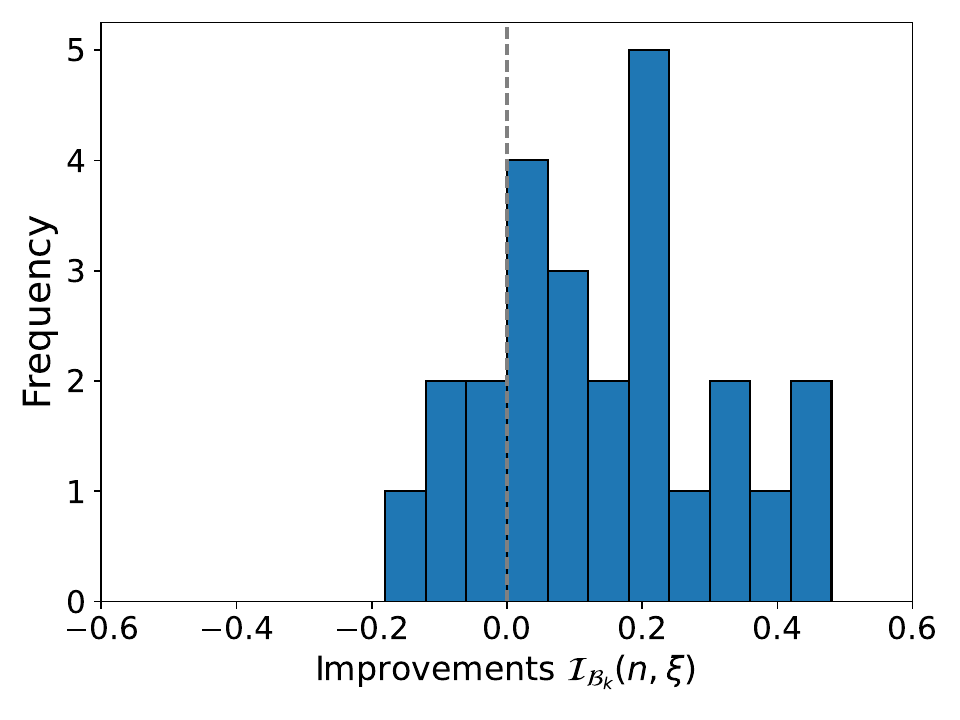}
        \caption{Tables}
        
    \end{subfigure}
    \hfill
    \begin{subfigure}{0.49\linewidth}
        \centering
        \includegraphics[width=\linewidth]{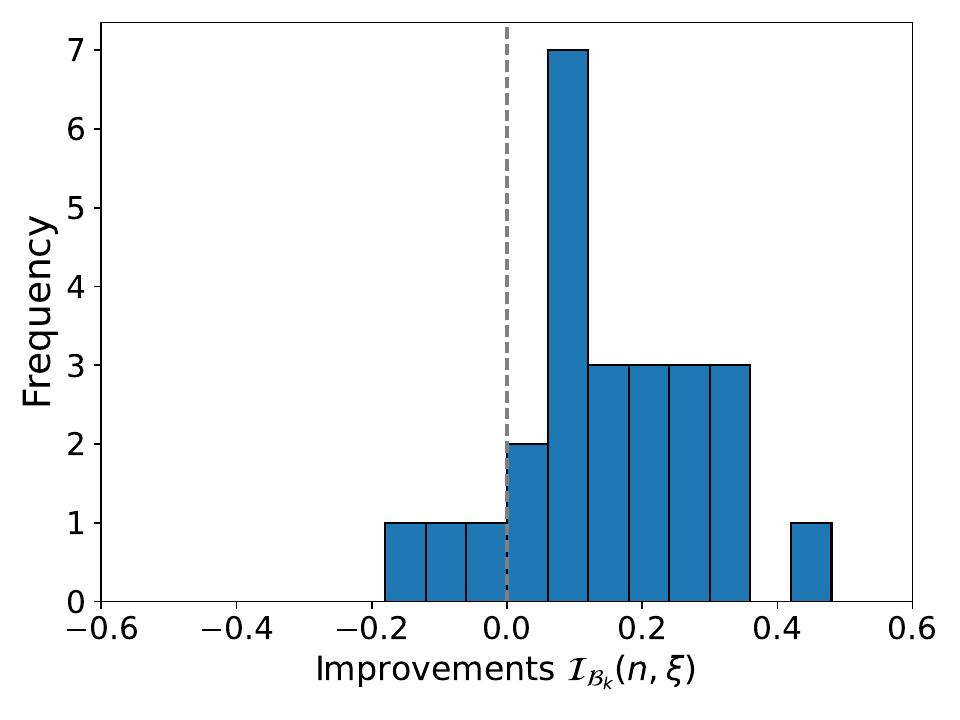}
        \caption{Airplanes}
        
    \end{subfigure}
    \hfill
    \caption{Improvement of the Neural UDF precision for ShapeNet surfaces.}
    \label{fig:histograms_of_improvements}
\end{figure}

Histograms of the improvements are plotted in Figure \ref{fig:histograms_of_improvements}. For this experiment, we set $n=600$ and $\xi=0.6$. Looking carefully at the charts, we observe that for some shapes our method does not improve the precision of Neural UDFs. However, this precision is improved for 76\% of the chairs, 72\% of the cars, 80\% of the tables and 88\% of the airplanes. The average improvement is around 15\% for the four categories. 

\section{Conclusion}

In this paper, we propose a new statistical method for edge detection on unstructured point clouds. This method is based on two key ideas. First, we use locally a statistical test of symmetry. The idea is to quantify the position of points with respect to their nearest neighbors. Second, we use a generalization of the notion of average to center circular data. This makes the circular data independent from any choice of starting point on the circle. Given observations distributed on the unit circle, it allows to compute a central symmetry test $p$-value that does not depend on any reference axis for polar coordinates. This allowed us to build our method on a basic goodness-of-fit test: Kolomorog-Smirnov's test.

Our edge detection method shows better results than commonly used geometric descriptors on surfaces of varying complexities. Moreover, this method is original since it is based on a local statistical test of central symmetry. This idea could inspire new hybrid approaches. On the one hand, Pauly's descriptor does not detect sharpest edges, which are well captured by Kolmogorov-Smirnov's descriptor. On the other hand, Kolomogorov-Smirnov's descriptor does not capture obtuse edges, which can be detected by Pauly's descriptor. Further work will be done to combine both descriptors and achieve better accuracy in edge detection. 

We show that this method can be used to improve the learning of 3D object representations. In particular, when learning the distance functions of 3D surfaces using neural networks, the training effort can be concentrated around surface edges. This paper explains in details the sampling procedure used to build training datasets that are denser around surface edges. This sampling procedure can be used for any application in which one needs to concentrate the training effort of a machine learning model in some areas of interest in the input space.

We propose a method to measure the influence of our edge detection method on the trained neural UDF accuracy. In particular, we use an optimization method that can be used for any application in which one needs to compute the iso-level set of a differentiable function.

On the one hand, our edge detection method improves the reconstruction capacity of neural networks by around 15\% on the ShapeNet dataset. On the other hand, when training distance functions with a limited number of training points, oversampling can achieve satisfactory accuracy with fewer training points. Therefore, we show that the use of edge detection allows to learn more accurately the shape UDFs. As a result, we obtain a more expressive representation of the shapes with a more efficient procedure. 

However, our edge detection method is subject to scaling and decision parameters that need to be determined beforehand for each dataset. A multiscale approach and a more granular classification of points according to edge sharpness will be performed in further work.

Our method improves the accuracy of Neural UDFs locally, close to surface edges. This improves the worst areas, since the reconstruction error measured in Hausdorff distance is reduced. However, when the error is averaged over the entire surface (e.g. Chamfer or Wasserstein distances), it is not improved in a consistent way all over the surfaces. To further improve the method, we will implement the calculation of a trade-off between edge and non-edge sampling.

Also, in this paper we show that our novel statistical method of surface edge detection allows to improve the accuracy of Neural UDFs. The latter are simple neural networks trained to fit the surface UDFs separately. In further work, we will use our statistical descriptor to improve the accuracy of more complex models like DeepSDF, that learn the UDFs of a whole set of surfaces in one single neural network. This will allow to build lower-dimensional representations of 3D shapes and potentially yield better performance in many geometric machine learning problems. 

\section*{Acknowledgements}

We are grateful to Sebastien Da Veiga, Xavier Roynard, Brian Staber, Thomas Pellegrini, Mathis Deronzier, Lucas De Lara, Simon Bartels for the fruitful discussions. Our work has been motivated and supported by Safran Tech company. They provided us with a dataset containing 3D turbine blade meshes and the results of numerical simulations. This data was crucial for running our experiments and build the intuitions that led to our contributions. This work has been done with the support of the Artificial and Natural Intelligence Toulouse Institute (ANITI).


\bibliographystyle{alpha}
\bibliography{refs}

\newpage


\appendix

    \section{Comparing Our Local Descriptor with Pauly's on 3D Toy Problems}
	\label{sup_mat:compare_ks_weinmann}

In this Section, we compare our edge detection method (called Kolmogorov-Smirnov descriptor in reference to the test statistic used) with the standard geometric descriptor proposed by Pauly et al. in \cite{pauly2002efficient}. This geometric descriptor is defined in Section \ref{subsection:intro_edge_detection}. It is hereafter referred to as Pauly's descriptor. For our comparison, we generated toy data illustrating all types of surface edges. This toy data is described in Section \ref{subsection:3d_toy_problem}. The results obtained with Kolmogorov-Smirnov's and Pauly's descriptors are compared in Section \ref{subsection:comparison_of_descriptors}.

		\subsection{3D toy reconstruction problem}
		\label{subsection:3d_toy_problem}

\begin{figure}[b]
    \centering
    \includegraphics[width=.8\linewidth]{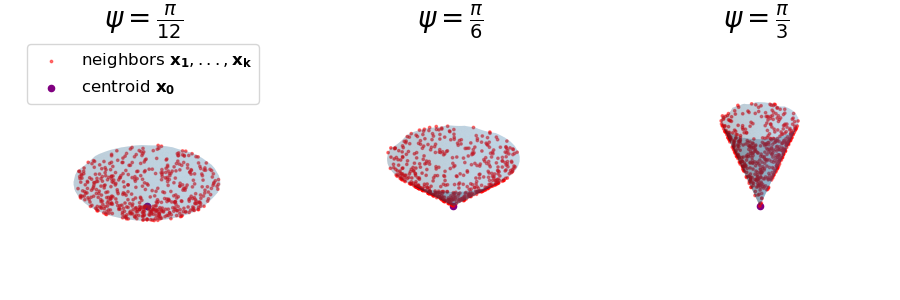}
    \caption{Different cases for local spikes. The parameter $\psi$ represents the rotation angle. When $\psi=0$, the cone is fully open. The higher $\psi$ the more closed the cone. These plots were generated with 500 samples each.}
    \label{fig:toy_cones}
\end{figure}

\begin{figure}
    \centering
    \includegraphics[width=.8\linewidth]{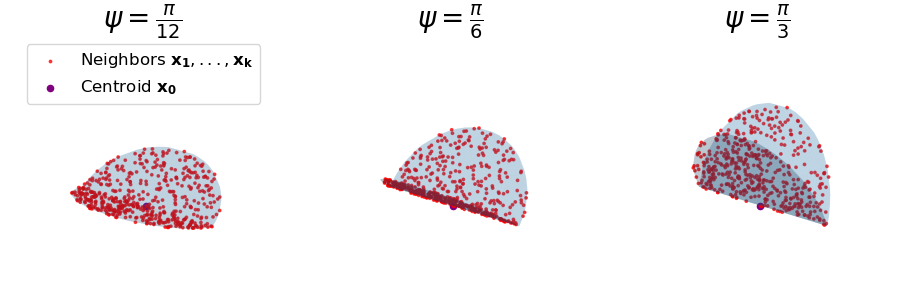}
    \caption{Different cases for local folds. The parameter $\psi$ represents the rotation angle. When $\psi=0$, the fold is fully plane. The higher $\psi$ the more folded the fold. These plots were generated with 500 samples each.}
    \label{fig:toy_folds}
\end{figure}

To reproduce the different types of surface edges commonly found on 3D objects, we generate surface portions corresponding to cones (Figure \ref{fig:toy_cones}) and folds (Figure \ref{fig:toy_folds}). 

\begin{figure}
    \centering
    \includegraphics[width=.8\linewidth]{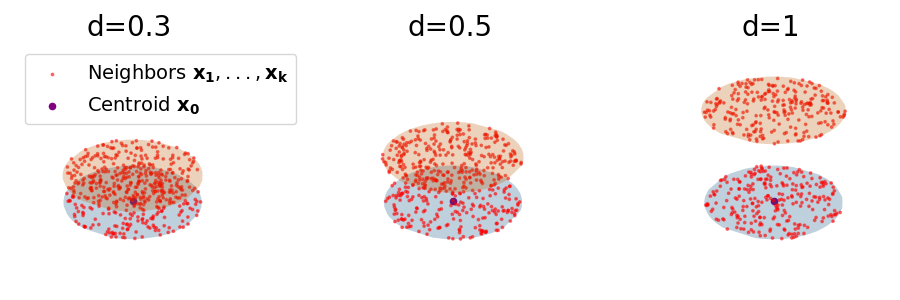}
    \caption{Different cases for fine plates. The parameter $d$ represents the thickness of the plate. The plots were generated with 500 points: each face of the plate is sampled with 250 points from the uniform distribution on unit circle.}
    \label{fig:toy_plates}
\end{figure}

It turns out that with Pauly's method, areas corresponding to thin plates (such as the back of a chair or the top of a table, for example) are erroneously detected as surface edges. We have therefore also generated surface portions corresponding to thin plates (Figure \ref{fig:toy_plates}).

For the cones, we begin by uniformly sampling points on a disk in $xy$ plane, which we denote as $\bm{S_0}$. For each point $\bm{x} \in \bm{S_0}$, we define the radial plane of $\bm{x}$ as the plane containing the $z$-axis and the radial axis of $\bm{x}$. We define the tangent axis $T(\bm{x})$ as the axis that contains the centroid point $\bm{x_0}$ and that is orthogonal to the radial plane of $\bm{x}$. Next, for each angle $\psi$ between $0$ and $\frac{\pi}{2}$, we can define a new set of points $S(\psi)$ as the set of the images of each point $\bm{x}$ in $\bm{S_0}$ by the rotation of angle $\psi$ around the tangent axis $T(\bm{x})$. This allows to generate a series of surfaces $S(\psi)$ ($\psi$ between $0$ and $\frac{\pi}{2}$) that can be thought of as images of a cone gradually closing. The centroid point for all the surfaces is the origin of the reference frame $xyz$. We define the range of values of $\psi$ such that $\psi=0$ corresponds to the initial disk (an fully open cone: $S(\psi=0)=\bm{S_0}$), and $\psi=\frac{\pi}{2}$ corresponds to a set of points aligned on the $z$-axis (a fully closed cone). On Figure \ref{fig:toy_cones}, cones are depicted for three values of $\psi$.

For the folds, we start with the same set of points $\bm{S_0}$ in $xy$ plane. However, here we rotate each point around the $y$-axis. The resulting series of surfaces can be seen as pictures of a fold that is gradually folded in half. On Figure \ref{fig:toy_folds}, folds are depicted for three values of $\psi$.

For the thin plates, 500 points are sampled uniformly on the unit disk in 2D. These points are assumed to lie in the plane $(z=0)$. We then repeat the same operation, except that the 500 new points obtained are considered to be in the plane $(z=d)$ (where $d$ - the parameter of interest - is the thickness of the platter). The centroid point for the local descriptor is the origin (0,0,0). On Figure \ref{fig:toy_plates}, thin plates are depicted for three values of $d$.

		\subsection{Comparison of the descriptors}
		\label{subsection:comparison_of_descriptors}

In this Section, we compare the values of Pauly's and Kolmogorov-Smirnov's local descriptors on toy surface portions (the cones, folds and thin plates described in Section \ref{subsection:3d_toy_problem}). The aim here is to assess the ability of each of the descriptors to detect surface edges.

To compute the descriptor over the entire surface, we must first select the neighborhood of each point on the surface. Neighborhood selection is formalized in Section \ref{subsection:statistical_edge_detection}.

\begin{figure}
    \begin{subfigure}{.32\linewidth}
  		\centering
  		\includegraphics[width=.99\linewidth]{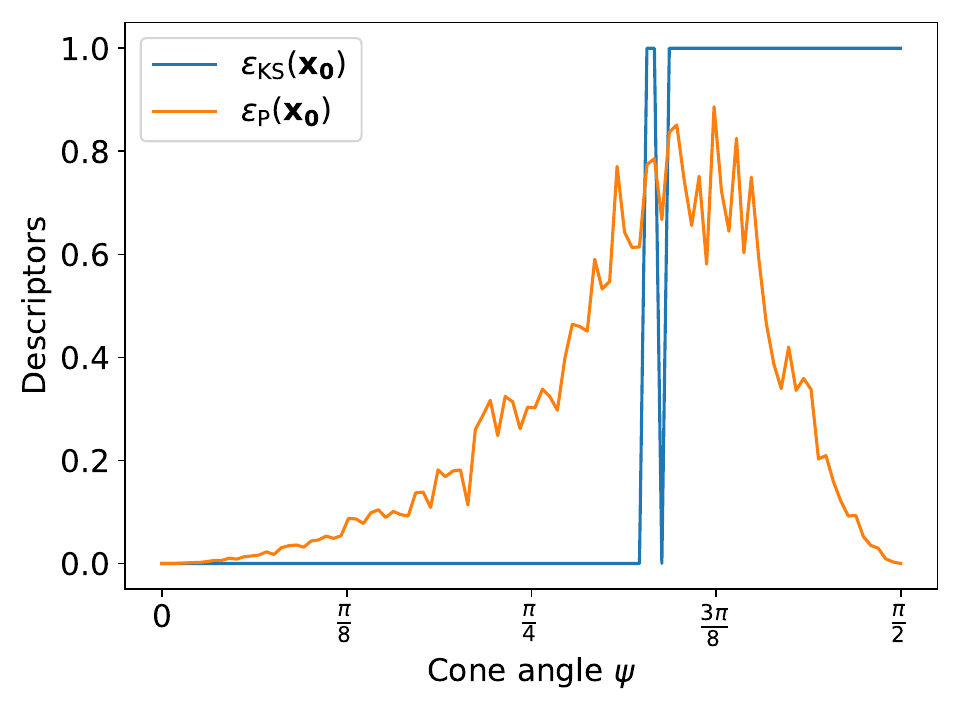}
  		\caption{Cones}
  		\label{subfig:results_toy_cones}
	\end{subfigure}
	\begin{subfigure}{.32\linewidth}
  		\centering
  		\includegraphics[width=.99\linewidth]{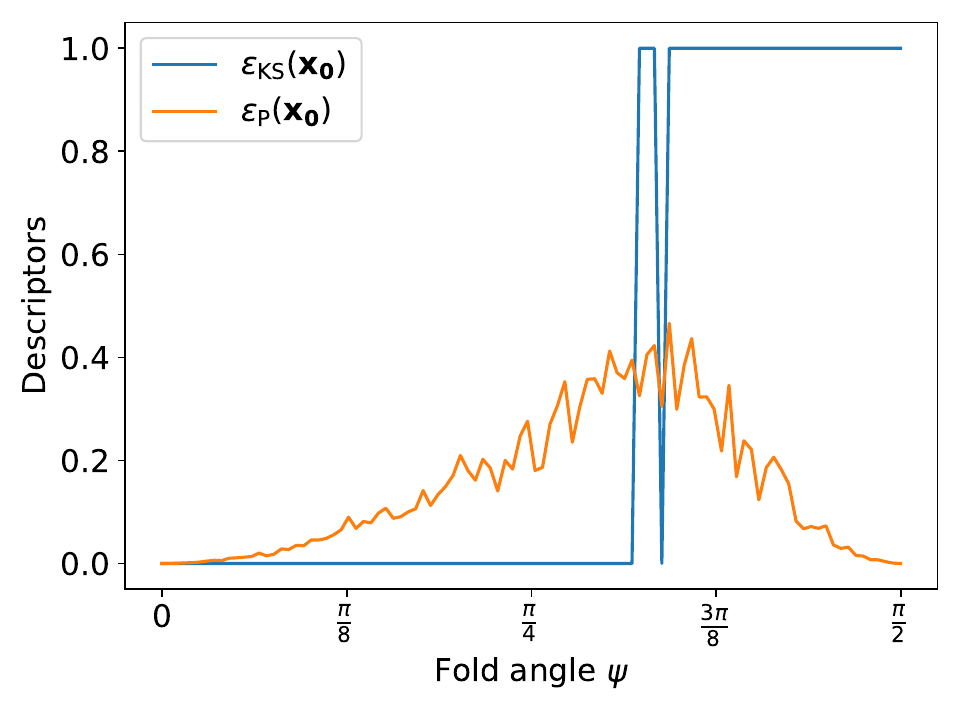}
  		\caption{Folds}
  		\label{subfig:results_toy_folds}
	\end{subfigure}
	\begin{subfigure}{.32\linewidth}
  		\centering
  		\includegraphics[width=.99\linewidth]{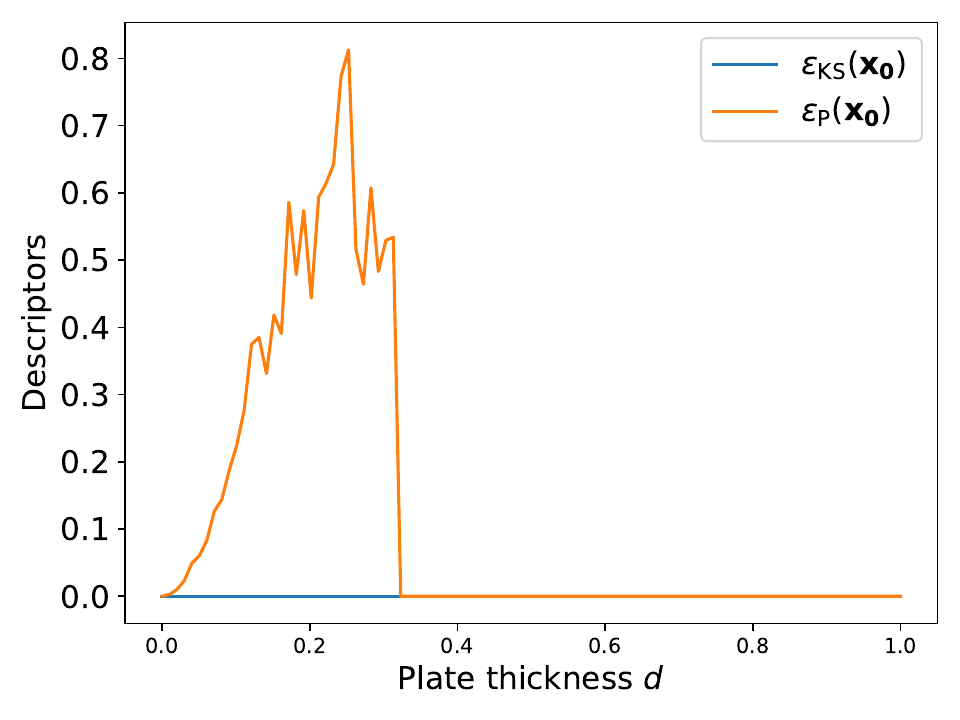}
  		\caption{Thin plates}
  		\label{subfig:results_toy_plates}
	\end{subfigure}
    \caption{Figures \ref{subfig:results_toy_cones} and \ref{subfig:results_toy_folds}: Estimator values for cones and folds with different angles (descriptor values are normalized between 0 and 1). When $\psi=0$, the cone (\textit{resp.} fold) is fully open and the surface is locally planar and both Pauly and Kolmogorov-Smirnov descriptors are equal to zero. The discontinuity on the Kolmogorov-Smirnov curve indicates a change of principal plane. For lower values of theta, the principal plane is tangent to the surface, while for higher values of theta, it is orthogonal to the surface. Pauly's descriptor is not suitable because for large values of theta it is not increasing, whereas the larger theta, the more pointed the cone (and therefore we would expect the descriptor to be larger). Figure \ref{subfig:results_toy_plates}: For thin plates ($d<0.3$), some points on the other side of the plateau are included in the neighborhood and this induces an error in edge detection: Pauly's descriptor appears to be very sensitive to this artifact, taking on very large values despite the fact that the surface is completely flat. Whereas Kolmogorov-Smirnov's descriptor is robust to this difficulty (its values remain low whatever the thickness of the plateau).}
    \label{fig:results_for_3d_toy}
\end{figure}

Figure \ref{fig:results_for_3d_toy} supports the conclusion that Pauly's descriptor is not effective in detecting all the edges. Specifically, for large values of $\psi$, Pauly's descriptor decreases as the local sharpness of the surface increases. This descriptor measures the local variation of the surface and is smaller for flatter surfaces. However, for very sharp or locally folded surfaces, Pauly's descriptor also approaches zero. These areas represent sharp edges and should be detected as such. They correspond to folds with an angle close to $\frac{\pi}{2}$. Kolmogorov-Smirnov's descriptor is a better alternative as it can clearly differentiate between locally quasi-planar surfaces and sharp edges.

	\section{Building an Intuition in 2D}
	\label{sup_mat:intuition_in_2d}

In this Section, we describe the thought process that led to our statistical edge detection method. In particular, the first 2D version of the method is described. Here, surfaces are replaced by contours, and the aim is to detect contour edges. In Section \ref{subsection:toy_problem_2d}, we describe some synthetic data corresponding to contour portions. In Section \ref{subsection:odds_ratio_descriptor}, we explain our 2D edge detector based on a test of odds.

		\subsection{Toy problem in 2D}
		\label{subsection:toy_problem_2d}

\begin{figure}
    \centering
    \includegraphics[width=.8\linewidth]{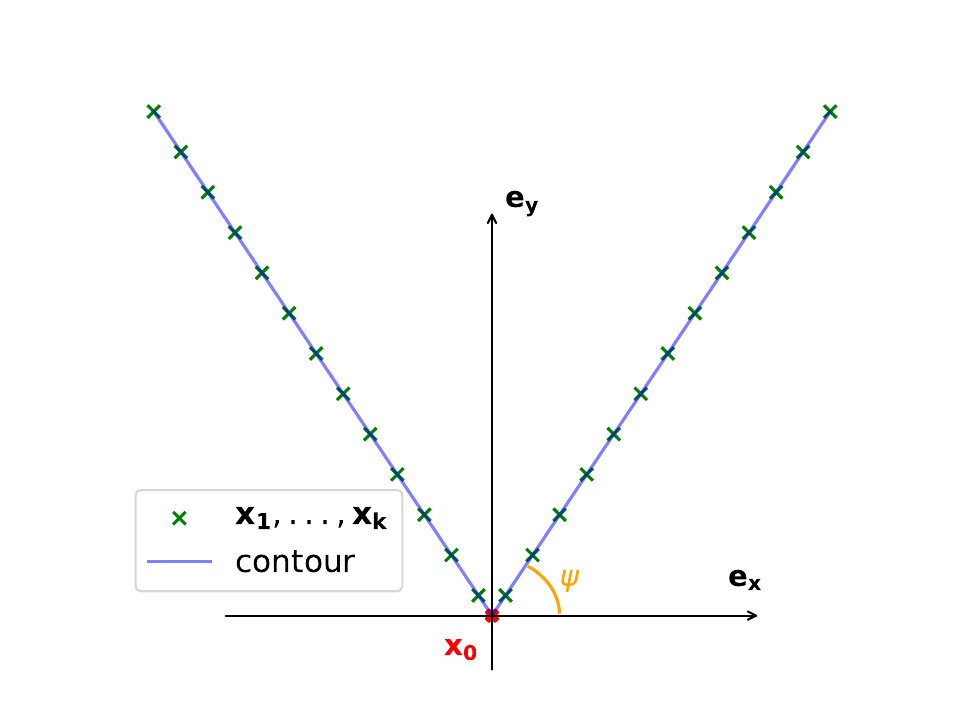}
    \caption{Illustration of the procedure used to generate the contour portions.}
    \label{fig:toy_problems_2d}
\end{figure}

We aim to compare different local descriptors for edge detection. Hence we generated contour edges with different angles. More precisely, we sample $k=50$ points evenly distributed on a straight line, along axis $\bm{e_x}$ in $\R^2$. Their $x$-coordinates are uniformly distributed between -1 and 1. Their $y$-coordinates are equal to zero. The obtained point cloud is denoted $\bm{C_0}$ corresponds to a straight contour portion. The centroid point is the origin of the Cartesian system. For any $\psi \in [0,\frac{\pi}{2})$, we apply a rotation to each point in $\bm{C_0}$ and generate the new contour portion 
\begin{equation}
    \bm{C_\psi} = \{ \Rc(\text{sgn}(\bm{x} \cdot \bm{e_x}) \psi) \bm{x}, \ \bm{x}\in\bm{C_0} \}
    \label{eq:define_contour_portion_psi}
\end{equation}
where
\begin{equation}
    \forall \psi \in [-\pi,\pi), \Rc(\psi) =
    \begin{pmatrix}
        \cos{\psi} & -\sin{\psi} \\
        \sin{\psi} & \cos{\psi}
    \end{pmatrix}.
\end{equation}
Figure \ref{fig:toy_problems_2d} illustrates how we generated the aforementioned contours portions.

		\subsection{Odds-ratio descriptor} 
		\label{subsection:odds_ratio_descriptor}

In this Paragraph, we describe the local statistical descriptor that we use to detect edges on 2D contours. We call it the \textit{odds-ratio descriptor}. In Section \ref{subsection:statistical_edge_detection}, we define the Kolomogorov-Smirnov descriptor as a novel statistical descriptor used to detect edges on 3D surfaces. This descriptor can be conceptualized as a 3-dimensional generalization of the odds-ratio descriptor defined here.

Let us explain the intuition that led us to the odds-ratio descriptor. We illustrate our intuition on the toy contour portions described in Paragraph \ref{subsection:toy_problem_2d}. These contour portions are depicted in Figure \ref{fig:toy_problems_2d}. A good descriptor is expected to discriminate obtuse contour portions (small values of $\psi$) and acute contour portions (large values of $\psi$).

The odds-ratio descriptor relies on a statistical approach to differentiate between obtuse (small $\psi$) and acute (large $\psi$) contour edges. This approach is based on two key ideas: (1) projecting the neighboring points on their \textit{average axis}. Then (2) modeling the projections as observations of a probability distribution and assess the symmetry of the latter around the centroid point. Indeed, for obtuse edges (small $\psi$), the average axis is tangent to the contour, and the projections of the points on this axis are evenly distributed around the centroid. Whereas for acute edges (large $\psi$), the average axis is orthogonal to the contour, and the centroid is off-centered with respect to the projections of its neighbors.

In order to formalize the method, let introduce $\bm{x_0}$ a centroid point and $\bm{x_1},...,\bm{x_k}$ its $k$-nearest-neighbors.

The average axis of the neighboring points is defined as the first eigen axis of their covariance matrix
\begin{equation}
\bm{\mathcal{V}}\vcentcolon=\frac{1}{k+1}\sum_{i=0}^k (\bm{x_i}-\overline{\bm{x}})(\bm{x_i}-\overline{\bm{x}})^T
\label{eq:cov_mat_2_2}
\end{equation}
where 
\begin{equation}
\overline{\bm{x}} \vcentcolon= \frac{1}{k+1}\sum_{i=0}^k \bm{x_i}
\label{eq:barycenter_2_2}
\end{equation}
is the neighborhood's mean.

For the sake of formalization, we call $\bm{e_1}$ the first eigenvector of $\Vc$. Its direction is the average axis of the neighboring points.

We project the neighboring points on their average axis and center them with respect to the centroid point, that is:
\begin{equation}
    \forall i \in \llbracket 1,k \rrbracket, \ X_i \vcentcolon= (\bm{x_i}-\bm{x_0}) \cdot \bm{e_1} 
\end{equation}

We aim to assess if the scalar values $X_1,...,X_k$ are evenly distributed around 0. In order to do so, we consider these values as i.i.d. realizations of a random variable $X$ and assess the symmtery of $X$ around 0. It amounts in testing the null hypothesis
\[
    H_0: \text{"the distribution of $X$ is symmetric around 0"}.
\]

Let introduce the number of positive realizations:
\begin{equation}
    k_+ = \sum_{i=1}^{k} \mathbb{1}_{X_i \geq 0}
    \label{eq:k_plus}
\end{equation}

It can be shown that under the null hypothesis $H_0$, the quantity
\begin{equation}
V_k \vcentcolon= \frac{2k_+-k}{\sqrt{k}}
\label{eq:odds_ratio_statistics}
\end{equation}
can be approximated by a Standard Gaussian distribution.

The $p$-value is then computed as the probability of observing a test statistic at least as extreme as the one observed here or more extreme, given the null hypothesis. Specifically, if we denote by $F$ the cumulative distribution function of the Standard Gaussian distribution, then
\begin{equation}
\text{$p$-value}\vcentcolon=2\times(1-F^{-1}(|v|))
\label{eq:p_value_2_2}
\end{equation}
where $v$ is the observed value of the test statistic $V_k$.

In order to validate or reject the null hypothesis, we use a decision threshold $p_0$. Here we empirically set $p_0=0.01$. In particular, if the $p$-value is lower than $p_0$, we reject the null hypothesis, which means that the distribution of $X$ is not symmetric. In this case, we consider that the contour edge is acute. On the contrary, the $p$-value is higher than $p_0$, we validate the null hypothesis, which means that the distribution of $X$ is symmetric. In this case, we consider that the contour edge is obtuse \textit{i.e.} that the contour is planar or quasi-planar. We define the odds-ratio descriptor as follows:

\begin{equation}
\epsilon_{OR}(\bm{x_0}) \vcentcolon= 
\begin{cases}
1 \: \text{if \textit{$p$-value} $\leq p_0$.} \\
0 \: \text{if \textit{$p$-value} $> p_0$.}
\end{cases}
\label{eq:thresholded_descriptor_odds}
\end{equation}

\begin{figure}
    \centering
    \includegraphics[width=.8\linewidth]{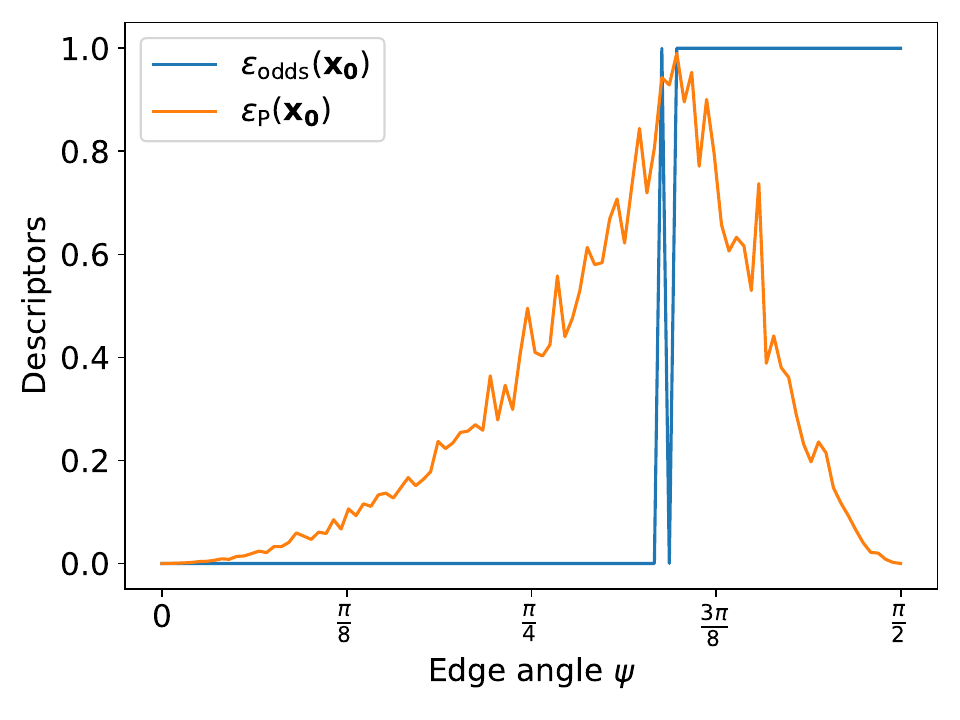}
    \caption{Descriptors computed on 2D toy contour portions as a function of the contour edge angle $\psi$. For 100 values of $\psi$ between 0 and $\frac{\pi}{2}$, we generate the contour portion $\bm{C_\psi}$ as defined in Equation \ref{eq:define_contour_portion_psi}. For each contour portion, we compute odds-ratio and Pauly's descriptors. It can be seen that Pauly's descriptor does not allow to discriminate between quasi-planar and acute contours. Indeed, it takes small values in both extreme cases ($\psi \simeq 0$ and $\psi \simeq \frac{\pi}{2}$).}
    \label{fig:results_toy_problems_2d}
\end{figure}

This method thus allows to remedy the two limitations of Pauly's approach:
\begin{enumerate}
\item This method clearly distinguishes situations 1) and 2) from situations 3) and 4).
\item The descriptor corresponds to the $p$-value of a statistical test, we can therefore interpret it and quantify its uncertainty as a function of the size of the neighborhood and the number of points sampled in it.
\end{enumerate}



\end{document}